\documentclass[runningheads]{llncs}

\usepackage{booktabs}
\usepackage{multirow}
\usepackage{graphicx}
\usepackage{amsmath}
\usepackage{amssymb}
\usepackage{float}
\usepackage[left]{lineno}
\usepackage{xcolor}
\usepackage[table]{xcolor}
\definecolor{evoshade}{RGB}{255,248,220}
\newif\ifanonymous
\anonymoustrue

\ifdefined \GramaCheck
  \newcommand{\CheckRmv}[1]{}
  \newcommand{\figref}[1]{Figure 1}%
  \newcommand{\tabref}[1]{Table 1}%
  \newcommand{\secref}[1]{Section 1}
  \renewcommand{\equref}[1]{Equation 1}
\else
  \newcommand{\CheckRmv}[1]{#1}
  \newcommand{\figref}[1]{Figure ~\ref{#1}}%

  \newcommand{\tabref}[1]{Table~\ref{#1}}%
  \newcommand{\secref}[1]{Section~\ref{#1}}
  \renewcommand{\eqref}[1]{Equation~(\ref{#1})}
\fi

\usepackage{ulem}

\DeclareMathOperator*{\argmax}{argmax}

 \anonymousfalse

\begin{document}

\title{Evolving Cache Schedules for Fast Diffusion Policy Inference}

\ifanonymous
  \author{Anonymous Author(s)}
  \authorrunning{Anonymous Author(s)}
  \institute{}
\else
  \author{
    Siying Wang\inst{1} \and
    Kangye Ji\inst{3} \and
    Di Wang\inst{2} \and
    Fei Cheng\inst{2,4}\thanks{Corresponding author.}
    }

    \authorrunning{S. Wang et al.}

    \institute{
    School of Telecommunications Engineering,
    Xidian University, Xi'an, China
    \and
    School of Computer Science and Technology,
    Xidian University, Xi'an, China
    \and
    Tsinghua Shenzhen International Graduate School, Tsinghua University, Shenzhen, China
    \and
    National Experimental Teaching Demonstration Center for Computer Network and Information Security, Xidian University, Xi'an, China\\
    \email{feicheng@xidian.edu.cn}
    }
   
\fi

\maketitle

\begin{abstract}
Diffusion policies achieve strong visuomotor control by iteratively denoising action chunks, but repeated denoising makes real-time deployment computationally demanding. Cache-based methods reduce inference cost by reusing intermediate activations, but existing training-free schedules typically allocate computation uniformly across blocks, ignoring heterogeneous redundancy across blocks and leading to a suboptimal performance--efficiency trade-off. To bridge this gap, we introduce Evolving Cache Schedules (EVO), a training-free acceleration framework that globally schedules cache refreshes via evolutionary search. EVO represents each candidate as a complete schedule over the block--timestep lattice. Thus, redundant transformer computations during iterative denoising can be skipped through cache reuse while preserving closed-loop rollout performance. To make the search practical, EVO introduces redundancy-aware initialization, which seeds the population with promising schedules, and target-conditioned early stopping, which verifies and terminates once a desired performance target is reached. The offline-optimized schedule can be directly plugged into pretrained diffusion policies without retraining. Extensive manipulation benchmarks show that EVO preserves near-full performance while substantially reducing computation, achieving up to $8.05\times$ action-generation speedup and reducing FLOPs from 15.77G to as low as 1.96G. Source code is available at \url{https://github.com/pillom/EVO}.

\keywords{Diffusion Policy \and Feature Caching \and Evolutionary Search}
\end{abstract}

\section{Introduction}

Diffusion policies have garnered substantial attention in robotic control for their ability to model multimodal action distributions through conditional denoising processes~\cite{chi2025diffusion,Ze-RSS-24,ho2020denoising,janner2022planning}. With scalable transformer denoisers, this formulation has become an expressive action-generation module for high-dimensional visuomotor policies and increasingly complex manipulation settings~\cite{peebles2023scalable,brohan2023rt,wang2025one}. However, the iterative denoising loop also imposes a heavy inference burden that directly limits the achievable action frequency, making it difficult for diffusion policies to satisfy the low-latency requirements of real-time, smooth robotic control~\cite{ji2026sparse}. Recently, cache-based acceleration methods provide a natural training-free approach to reducing the inference cost of diffusion policies. These methods exploit temporal
redundancy along the denoising trajectory by caching intermediate activations
and reusing them at skipped positions, so that only selected block--timestep
positions need to be recomputed~\cite{ma2024deepcache,lyu2025fastercache,liu2025faster,zhao2025real,zou2025accelerating}.

\begin{figure}[t]
    \centering
    \includegraphics[width=\linewidth]{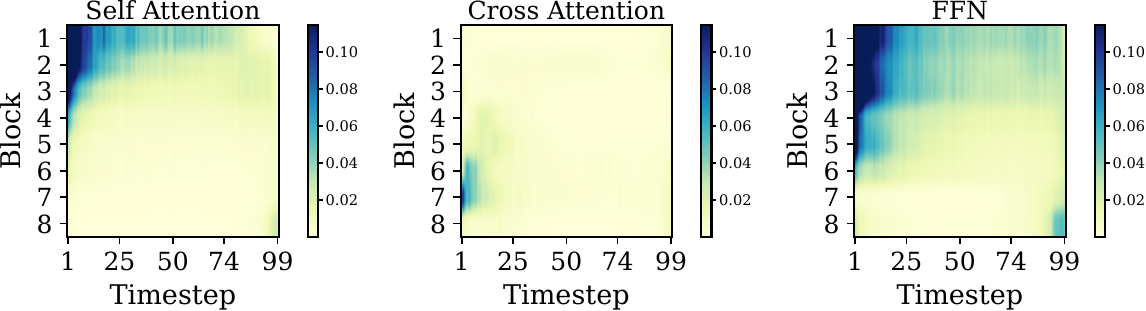}
\caption{
Cross-step feature dissimilarity of transformer modules on Push-T.
Higher dissimilarity indicates larger feature variation between denoising steps
and therefore weaker temporal redundancy. The heatmaps show that different
blocks exhibit distinct temporal distributions of high-dissimilarity regions
as well as different overall dissimilarity levels, revealing heterogeneous
redundancy patterns and degrees across blocks. FFN denotes the feed-forward network.
}
    \label{fig:block_step_variation}
\end{figure}

Existing training-free methods typically allocate computation across blocks uniformly, which ignores heterogeneous redundancy across blocks and leads to a suboptimal trade-off between performance and efficiency. EfficientVLA, for example, refreshes all blocks synchronously with a fixed temporal interval, while BAC chooses block-specific update timesteps but still assigns a fixed refresh budget to each block~\cite{yang2026efficientvla,ji2026blockwise}. To examine whether such uniform allocation is justified, we measure cross-step
feature dissimilarity across transformer blocks. Higher dissimilarity indicates
larger feature variation between denoising steps and therefore weaker computation redundancy. As shown in Fig.~\ref{fig:block_step_variation}, blocks differ not
only in the temporal distribution of high-dissimilarity regions, but also in
their overall dissimilarity levels, revealing distinct redundancy patterns and
degrees across blocks. 

This observation motivates a global cache scheduling strategy that allocates one fixed refresh budget over the entire block--timestep lattice, allowing computation to move from redundant blocks to sensitive ones. For concreteness, we view training-free acceleration as a fixed-budget cache-scheduling problem. Given cacheable blocks \(\mathcal{B}\), denoising steps \(\mathcal{T}\), and a budget \(K\), a schedule selects \(K\) block--step positions to refresh and reuses cached activations for the rest. A naive solution is exhaustive search, but enumerating \(K\) refreshes over the \(|\mathcal{B}||\mathcal{T}|\) block--step grid yields an exponential search space. Alternatively, one might extend similarity-based approximation~\cite{ji2026blockwise} to global scheduling, using feature similarity to estimate cache-induced errors without exhaustive evaluation. However, under coupled block schedules, inter-block error propagation makes update-induced errors difficult to formulate as a tractable optimization objective~\cite{ji2026blockwise}, limiting its applicability in global cache scheduling.

To bridge this gap, we propose EVO, a training-free framework that searches global cache schedules with an evolutionary algorithm. EVO represents each candidate as a complete refresh schedule and optimizes it through selection, crossover, mutation, and elitism, using closed-loop rollout performance as the fitness. While rollout-based fitness provides a direct estimate of deployment performance, its high computational cost limits the practicality of large-scale search. EVO therefore proposes two lightweight mechanisms to make search practical. Specifically, redundancy-aware initialization uses activation dissimilarity only as a soft prior, biasing the initial population toward less redundant block--step positions while keeping random individuals and full-space mutation for exploration. Meanwhile, target-conditioned early stopping turns search into a goal-driven procedure: once a quick evaluation finds a candidate that reaches a baseline-relative target, EVO confirms it with an independent formal evaluation before termination. The final schedule is then deployed as an offline-optimized cache plugin, without altering the policy parameters, diffusion sampler, or action interface.


In summary, our contributions are:
\begin{itemize}
    \item We reveal heterogeneous redundancy in Diffusion Policy that different transformer blocks exhibit distinct temporal redundancy patterns and degrees, motivating a global cache scheduling strategy.

    \item We propose EVO, a training-free evolutionary framework that searches for complete global cache schedules using closed-loop rollout performance as the optimization objective.

    \item We develop redundancy-aware initialization and target-conditioned early stopping to reduce the cost of rollout-driven evolutionary search while preserving global exploration.

    \item We validate EVO on multiple robotic manipulation benchmarks, showing that it preserves near-full policy performance while achieving up to 8.05\(\times\) action-generation speedup.
\end{itemize}
\section{Related Work}

\paragraph{Diffusion Policies.}  Diffusion
models were first popularised for image synthesis for their ability to
generate high-quality and diverse samples~\cite{ho2020denoising,rombach2022high}.
They were later introduced to visuomotor control because conditional denoising
provides a natural way to model multimodal action distributions~\cite{chi2025diffusion}. Diffusion Policy formulates action generation as conditional denoising and
supports both U-Net and transformer backbones~\cite{chi2025diffusion}. The transformer variant often
performs better on complex or high-action-rate tasks, but repeated transformer
computation across denoising steps makes inference latency more pronounced,
making it difficult to meet the low-latency requirements of real-time robotic
control~\cite{pmlr-v164-mandlekar22a}. Existing acceleration methods either reduce the number of denoising steps through faster samplers
~\cite{song2021denoising,lu2022dpm,lu2025dpm,zhao2023unipc}, or reduce computation through pruning and distillation~\cite{wang2025one,salimans2022progressive,ji2026sparse,pmlr-v202-song23a}. However, they typically alter the sampling process or require additional training. In contrast, our method accelerates transformer-based diffusion policies by reducing inference computation through caching, without retraining the model or modifying the original sampler.

\paragraph{Cache-based Acceleration.}
Feature caching stores and reuses intermediate features across denoising steps to avoid redundant computation. Existing methods have shown strong acceleration for image generation in both U-Net diffusion models and diffusion transformers~\cite{ma2024deepcache,wimbauer2024cache,selvaraju2024fora,zou2025accelerating,son2026relational,ma2024learning}. These methods typically rely on fixed refresh rules, token-wise or feature-wise update strategies, or learned routing networks, and they are mainly evaluated with image quality metrics such as Fréchet Inception Distance (FID). In action generation, EfficientVLA applies feature caching to the diffusion-based action head of Vision-Language-Action (VLA) models~\cite{yang2026efficientvla}, but its cache refresh still follows a fixed-interval rule and does not capture redundancy differences across blocks and denoising steps. Block-wise Adaptive Caching (BAC) further selects refresh times for each block based on feature similarity~\cite{ji2026blockwise}, but its refresh budget remains constrained within each block. Existing cache-based methods therefore still lack global budget allocation over the block--timestep lattice. Our method addresses this limitation by formulating cache scheduling as a global allocation problem and directly optimizing rollout success rate.

\paragraph{Evolutionary Search.}  Evolutionary algorithms and other black‑box optimisers are well suited to discrete
combinatorial problems where gradients are unavailable~\cite{zames1981genetic,real2019regularized,holland1992adaptation,stanley2002evolving}.  They have been
applied to policy search and neural architecture optimisation for robotic tasks~\cite{silva2015odneat,guo2025neural,hegde2023efficiently}, where candidate architectures or policies are often evaluated by empirical task performance. This setting is closely related to our cache scheduling problem: each schedule is a discrete subset of block--timestep refresh positions under a fixed budget, and its quality is determined by closed-loop rollout performance rather than a differentiable objective. We therefore use evolutionary search as an offline optimizer for complete cache schedules. Since black-box search with task-level evaluation often incurs high evaluation cost, EVO further incorporates redundancy-aware initialization and target-conditioned early stopping to reduce unnecessary evaluations.
\section{Method}
\label{sec:method}

\begin{figure}[t]
    \centering
    \includegraphics[width=\linewidth]{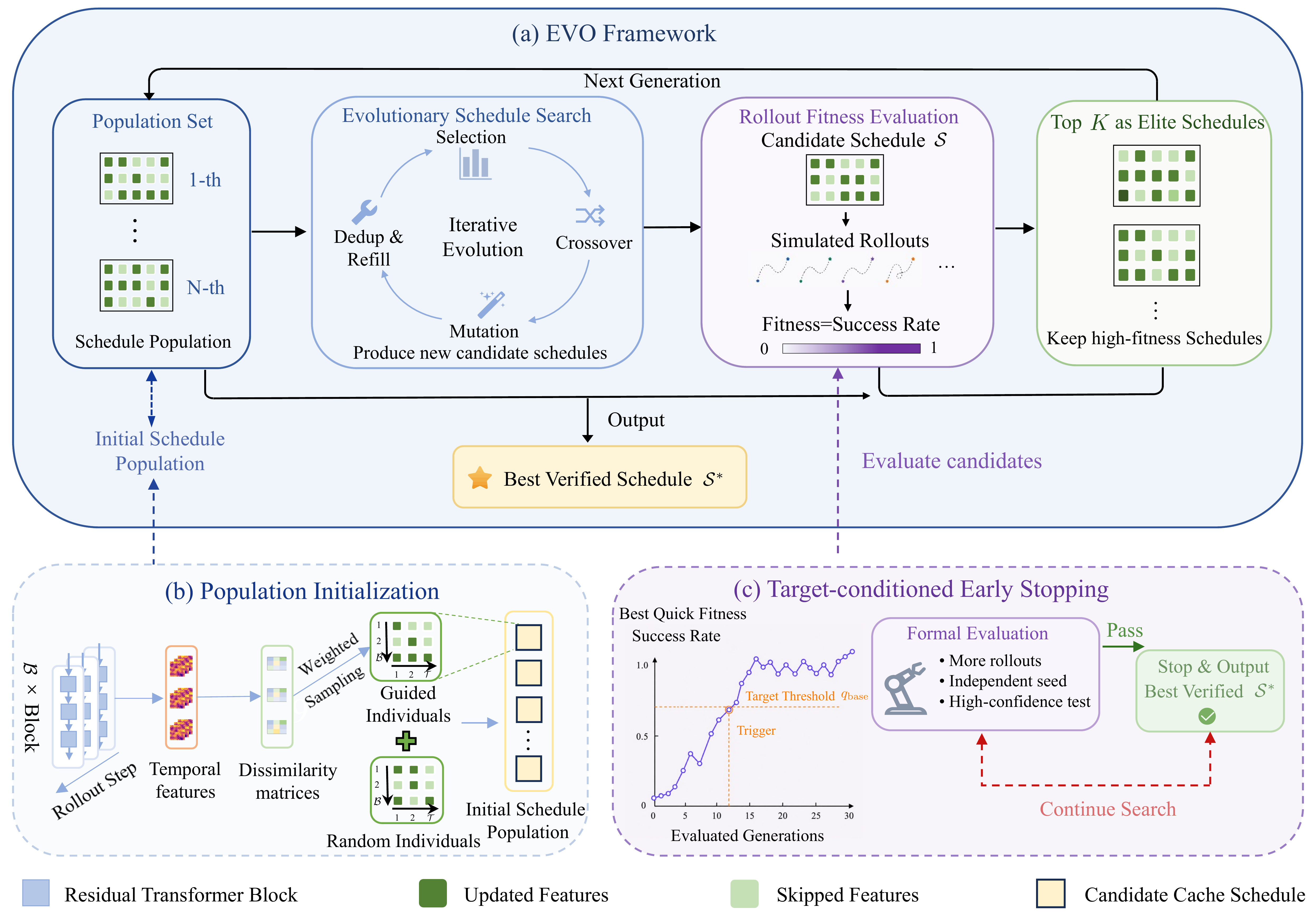}
    \caption{
    Framework of EVO, a training-free acceleration framework for global cache scheduling via evolutionary search.
    (a) EVO searches for a globally budgeted cache schedule, using rollout fitness as the selection criterion.
    (b) Redundancy-aware and random individuals are combined to initialize the schedule population.
    (c) Target-conditioned early stopping performs formal verification for promising schedules and terminates once the desired performance target is reached.
    }
    \label{fig:evo_framework}
\end{figure}

EVO accelerates a pretrained transformer-based diffusion policy by optimizing where cache refreshes are performed during iterative denoising. Rather than assigning a uniform or per-block refresh budget, EVO searches for a globally constrained subset of block--timestep positions under a fixed computation budget. As illustrated in Fig.~\ref{fig:evo_framework}, the method formulates cache scheduling as a subset-selection problem over the block--timestep lattice and optimizes complete schedules with evolutionary search, using closed-loop rollout performance as the fitness. Redundancy-aware initialization and target-conditioned early stopping are introduced to reduce the cost of this offline search. The resulting schedule is then utilized by the pretrained policy without retraining or modifying the diffusion sampler.

\subsection{Global Cache Scheduling}
\label{sec:global-cache-scheduling}

Let $\pi_\theta$ denote a pretrained diffusion policy. For each action query, the policy starts from initial noise and performs $T$ reverse denoising steps, indexed by $\mathcal{T}=\{0,\ldots,T-1\}$. At each denoising step, the transformer denoiser propagates the action representation through a sequence of cacheable computation units. For an $L$-layer transformer decoder, the self-attention, cross-attention, and feed-forward residual branches in each layer are used as cacheable units, yielding $B=3L$ units indexed by $\mathcal{B}=\{1,\ldots,B\}$. The cacheable computations of one action query can therefore be represented as a two-dimensional block--timestep lattice $\mathcal{B}\times\mathcal{T}$.

Each lattice position $(b,t)$ corresponds to the residual-branch computation of unit $b$ at denoising step $t$. EVO maintains one residual cache $c_b$ for each unit and resets all caches before the next action query. Let $r_{b,t}$ be the residual output of unit $b$ at denoising step $t$. 
Given a cache schedule $\mathcal{S}$, the residual $\tilde{r}_{b,t}$ used by the accelerated policy is
\begin{equation}
\tilde r_{b,t} =
\begin{cases}
r_{b,t}, & (b,t)\in \mathcal{S},\\
c_b, & (b,t)\notin \mathcal{S},
\end{cases}
\qquad
c_b \leftarrow r_{b,t} \ \text{if } (b,t)\in \mathcal{S}.
\label{eq:cache_execution}
\end{equation}
Thus, selected positions evaluate the original residual branch and refresh the cache, whereas unselected positions reuse the most recently cached residual of the same unit. Cache scheduling only changes which residual computations are executed,
while the policy parameters, observation encoder, diffusion sampler, and action interface remain unchanged.

\begin{figure}[!t]
    \centering
    \begin{minipage}[t]{0.48\linewidth}
        \centering
        \includegraphics[height=3.2cm]{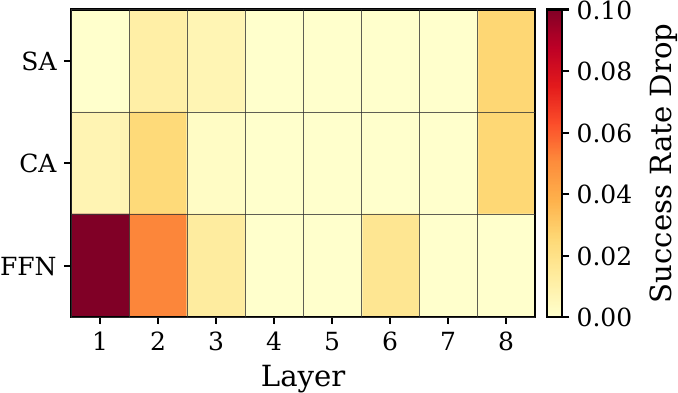}
        \vspace{-0.4em}
        
        \small (a) Single-block cache stress test
    \end{minipage}
    \hfill
    \begin{minipage}[t]{0.48\linewidth}
        \centering
        \includegraphics[height=3.2cm]{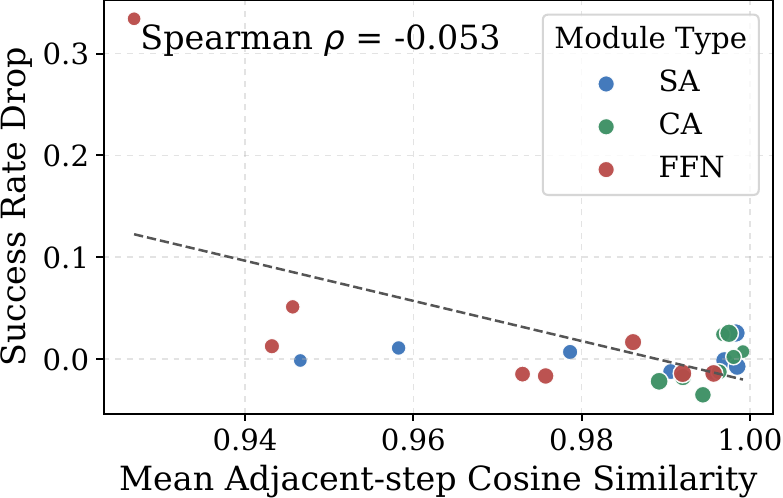}
        \vspace{-0.4em}
        \small (b) Feature similarity vs. task sensitivity
    \end{minipage}
    \caption{
    Motivation for rollout-driven global cache scheduling.
    (a) Under the same sparse update pattern, different transformer blocks cause different success drops, indicating non-uniform task sensitivity across layers and module types.
    (b) The weak correlation between adjacent-step feature similarity and success drop suggests that feature similarity is useful as a prior but insufficient as the final scheduling objective.
    }
    \label{fig:motivation}
\end{figure}

Per-block scheduling constrains different blocks to receive comparable refresh allocations, although their effects on closed-loop control may differ substantially. This assumption is evaluated by applying the same sparse update pattern to each block individually while keeping all other blocks fully computed. As shown in Fig.~\ref{fig:motivation}(a), the resulting success drop varies across blocks, indicating clear block-dependent task sensitivity. This motivates a global allocation strategy in which refreshes are assigned over the full lattice rather than fixed within each block.

Motivated by this non-uniformity, EVO defines a cache schedule as a fixed-budget subset of the block--timestep lattice:
\begin{equation}
    \mathcal{S} \subseteq \mathcal{B}\times\mathcal{T}, \qquad |\mathcal{S}|=K,
    \label{eq:evo_schedule_budget}
\end{equation}
where $K$ denotes the total refresh budget, i.e., the number of lattice positions at which the original residual computation is executed. Unlike per-block scheduling, this formulation does not pre-assign a fixed number of refreshes to each block, allowing the budget to be allocated globally according to downstream performance.

\subsection{Evolutionary Search over Schedules}
\label{sec:evolutionary-search}

The fixed-budget search space contains $\binom{\mathcal{B}\mathcal{T}}{K}$ feasible schedules, making exhaustive enumeration impractical. EVO therefore employs evolutionary search to optimize complete cache schedules in this large, discrete, and non-differentiable space. In the evolutionary algorithm, we represent each schedule $\mathcal{S}$ by an
individual $x$, which contains $K$ unique refresh positions:
\begin{equation}
x=\{(b_i,t_i)\}_{i=1}^{K},
\qquad
(b_i,t_i)\in\mathcal{B}\times\mathcal{T},
\qquad
|x|=K.
\label{eq:individual_encoding}
\end{equation}

Feature similarity is commonly used as a proxy objective for cache scheduling, based on the intuition that similar activations across adjacent denoising steps can be reused with limited error. In closed-loop control, however, activation similarity is not necessarily aligned with task-level sensitivity. Fig.~\ref{fig:motivation}(b) shows that blocks with comparable similarity scores can induce different success drops, and the overall correlation between feature similarity and task sensitivity is weak. Additional analysis across Kitchen, Square MH, Tool PH, Transport MH,
and Push-T shows a similar trend. Detailed results are provided in Sec.~A.4 of the supplementary material. EVO therefore uses feature similarity only for initialization and adopts rollout success rate as the fitness for schedule optimization.

Given a candidate schedule $\mathcal{S}$, EVO applies the corresponding cached execution to the pretrained policy, denoted by $\pi_\theta(\mathcal{S})$, and estimates its empirical success rate over evaluation episodes:
\begin{equation}
F(\mathcal{S})=
\frac{1}{N_{\mathrm{eval}}}
\sum_{n=1}^{N_{\mathrm{eval}}}
\mathbf{1}\{\mathrm{Success}(\tau_n;\mathcal{S})\},
\qquad
\tau_n\sim\pi_\theta(\mathcal{S}).
\label{eq:fitness}
\end{equation}
This fitness evaluates the effect of a cache schedule on the closed-loop behavior of the accelerated policy. EVO optimizes the schedule by maximizing the downstream rollout success rate under the fixed refresh budget:

\begin{equation}
\mathcal{S}^{*}
=
\argmax_{|\mathcal{S}|=K}
F(\mathcal{S}), \qquad \mathcal{S}\subseteq \mathcal{B}\times\mathcal{T}.
\label{eq:evo_objective2}
\end{equation}

The evolutionary procedure iteratively applies tournament selection, set-level crossover, mutation, repair, and elitism. Tournament selection samples parent schedules according to their rollout fitness. Set-level crossover exchanges refresh positions between two parents, while mutation resamples a subset of positions from the full block--timestep lattice. The repair operator removes duplicate positions and fills missing positions so that each child remains a feasible schedule with exactly $K$ unique refreshes. Elitism preserves the best-performing schedules across generations, and evaluated schedules are stored using canonical signatures to avoid repeated rollouts. These operators maintain the fixed-budget constraint while allowing refresh allocations to move freely across blocks and denoising steps.

\subsection{Practical Search Mechanisms}
\label{sec:practical-search}

Although rollout-based fitness directly reflects closed-loop control performance, it is expensive because each candidate must be evaluated through environment rollouts. EVO reduces this offline cost with two search-time mechanisms. Redundancy-aware initialization improves the starting population, and target-conditioned early stopping limits unnecessary rollout evaluations. Both mechanisms affect only the search process; schedule selection remains governed by rollout fitness.

\paragraph{Redundancy-aware Initialization.}
Feature similarity is not used as the final optimization objective, but it provides a useful prior for constructing the initial population. While uniformly random initialization provides broad exploration, it may allocate early refresh positions to highly redundant parts of the lattice. EVO therefore uses activation redundancy to bias part of the initial population toward less redundant block--timestep positions.

Residual activations are collected from a small set of rollout steps generated by the uncached policy. Let $a_{r,b,t}$ denote the activation for rollout index $r$, unit $b$, and denoising step $t$. Each block--timestep position is assigned the following dissimilarity score:
\begin{equation}
d_{b,t}
=
\mathbb{E}_{r}
\left[
\frac{1}{T-1}
\sum_{t'\ne t}
\left(
1-
\frac{
\langle a_{r,b,t}, a_{r,b,t'} \rangle
}{
\|a_{r,b,t}\|_2 \|a_{r,b,t'}\|_2
}
\right)
\right].
\label{eq:redundancy_score}
\end{equation}
A larger $d_{b,t}$ indicates that the activation at position $(b,t)$ is less similar to other denoising steps within the same module, and hence has lower feature redundancy. After normalizing $d_{b,t}$ into $w_{b,t}$, guided individuals sample refresh positions according to
\begin{equation}
p_{b,t}
=
\frac{(\epsilon+w_{b,t})^\gamma}
{
\sum_{(b',t')\in\mathcal{B}\times\mathcal{T}}
(\epsilon+w_{b',t'})^\gamma
}.
\label{eq:guided_sampling}
\end{equation}
This distribution assigns larger initial sampling probabilities to less redundant positions. The prior is used only for initialization: the initial population also contains random schedules, and subsequent mutation and repair can sample any position in $\mathcal{B}\times\mathcal{T}$. Rollout fitness remains the only selection criterion.

\paragraph{Target-conditioned Early Stopping.}
To avoid unnecessary evaluations after a satisfactory schedule has been found, EVO terminates search once a candidate reaches the prescribed performance target. Let $q_{\mathrm{base}}$ denote the formal rollout score of the uncached policy. During evolution, candidate schedules are first screened with a low-cost quick evaluation. When a schedule satisfies
\begin{equation}
q_{\mathrm{quick}}(\mathcal{S}) \ge q_{\mathrm{base}},
\label{eq:quick_trigger}
\end{equation}
EVO performs an independent formal evaluation using a larger rollout budget and independent random seeds. The schedule is accepted, and search terminates only if
\begin{equation}
q_{\mathrm{formal}}(\mathcal{S}) \ge q_{\mathrm{base}}-\delta_{\mathrm{acc}},
\label{eq:formal_accept}
\end{equation}
where $\delta_{\mathrm{acc}}$ is the allowed absolute performance drop. The quick evaluation therefore serves as a screening stage for promising candidates, while the formal evaluation provides the final acceptance criterion and mitigates the effect of noisy quick rollouts. Representative search trajectories in Sec.~A.6 of the supplementary material illustrate that this
mechanism reduces unnecessary evaluations while avoiding premature termination.

\subsection{Offline-Optimized Deployment}
\label{sec:deployment}

After search, EVO fixes the best verified schedule and removes the optimizer from the inference loop. At test time, the deployment wrapper loads $\mathcal{S}^{*}$, resets the residual caches at the beginning of each action query, and follows Eq.~\ref{eq:cache_execution} to determine whether each block--timestep position is refreshed or replaced by cache reuse. The policy weights, observation encoder, diffusion sampler, and action interface remain unchanged. The search is performed offline only once for a given task. The resulting schedule is reused across subsequent episodes drawn from the same environment distribution, without online adaptation or repeated search. Further analysis is provided in Sec.~\ref{sec:ablstudy}.
\section{Experiments}

This section evaluates whether EVO can reduce diffusion-policy inference cost
while preserving closed-loop control performance, without retraining or modifying
the pretrained policy. We first introduce the experimental setup, including
benchmark tasks, policy backbones, evaluation metrics, and implementation details.
We then compare EVO with representative training-free acceleration baselines to
evaluate the trade-off between task success rate and inference speed. Finally, we
conduct ablation studies to analyze how global block-step scheduling and
similarity-guided initialization contribute to the overall performance.

\begin{table}[H]
\centering
\caption{Benchmark on Proficient Human (PH) demonstration data. Success rates are evaluated by mean $\pm$ standard deviation over three evaluation seeds, and speedups are measured with full DP-T inference. EVO achieves a favorable trade-off between task success rate and inference speed, maintaining an average success rate of 0.78 while reaching up to $8.05\times$ speedup.}
\label{tab:ph_results}
\setlength{\tabcolsep}{4pt}
\resizebox{\textwidth}{!}{
\begin{tabular}{lccccccccc}
\toprule
\multirow{2}{*}{Method}
& \multicolumn{6}{c}{Success Rate $\uparrow$}
& \multirow{2}{*}{AVG}
& \multirow{2}{*}{FLOPs}
& \multirow{2}{*}{Speed $\times$} \\
\cmidrule(lr){2-7}
& Lift & Can & Square & Transport & Tool & Push-T & & & \\
\midrule
Full Precision
& $\mathbf{1.00} \pm 0.00$
& $\mathbf{0.99} \pm 0.01$
& $0.85 \pm 0.04$
& $\mathbf{0.82} \pm 0.00$
& $0.45 \pm 0.05$
& $0.66 \pm 0.02$
& \textbf{0.80} & 15.77G & -- \\
\midrule
EfficientVLA  ($M=8$)
& $\mathbf{1.00} \pm 0.00$
& $0.79 \pm 0.06$
& $0.85 \pm 0.05$
& $0.69 \pm 0.07$
& $0.25 \pm 0.10$
& $0.64 \pm 0.02$
& 0.70 & 1.64G & 9.62 \\
EfficientVLA ($M=10$)
& $0.89 \pm 0.05$
& $0.10 \pm 0.05$
& $0.40 \pm 0.04$
& $0.23 \pm 0.05$
& $0.00 \pm 0.00$
& $0.59 \pm 0.03$
& 0.37 & 1.95G & 8.09 \\
\midrule
BAC ($M=8$)
& $\mathbf{1.00} \pm 0.00$
& $0.63 \pm 0.03$
& $0.86 \pm 0.05$
& $0.77 \pm 0.08$
& $0.37 \pm 0.09$
& $0.66 \pm 0.03$
& 0.72 & 2.33G & 6.77 \\
BAC ($M=10$)
& $\mathbf{1.00} \pm 0.00$
& $0.63 \pm 0.02$
& $\mathbf{0.91} \pm 0.05$
& $\mathbf{0.81} \pm 0.04$
& $\mathbf{0.48} \pm 0.09$
& $0.66 \pm 0.01$
& 0.75 & 2.72G & 5.80 \\
\midrule
\rowcolor{evoshade}
EVO ($M=8$)
& $\mathbf{1.00} \pm 0.00$
& $\mathbf{0.98} \pm 0.04$
& $0.88 \pm 0.02$
& $0.74 \pm 0.02$
& $0.39 \pm 0.08$
& $0.66 \pm 0.03$
& \textbf{0.78} & 1.96G & 8.05 \\
\rowcolor{evoshade}
EVO ($M=10$)
& $\mathbf{1.00} \pm 0.00$
& $0.97 \pm 0.01$
& $0.87 \pm 0.02$
& $0.76 \pm 0.04$
& $0.41 \pm 0.05$
& $\mathbf{0.69} \pm 0.03$
& \textbf{0.78} & 2.33G & 6.77 \\
\bottomrule
\end{tabular}
}
\end{table}

\subsection{Experimental Setup}

\noindent\textbf{Models, Benchmarks, and Metrics.}
Following the standard setting of Diffusion Policy, we adopt the Transformer-based Diffusion Policy (DP-T) as the base policy and evaluate EVO by publicly released pretrained checkpoints. We conduct experiments on multiple robotic manipulation benchmarks, including RoboMimic tasks, Push-T, Block Push, and Kitchen. For the multi-stage results in Table~\ref{tab:multi_stage_results}, BP and Kit
denote Block Push and Kitchen, respectively, and AVG denotes the average
over all reported success-rate metrics. The demonstration data comprises mixed proficient/non-proficient human (MH) teleoperation demonstrations, proficient human (PH) teleoperation demonstrations, and expert trajectories generated by scripted Markovian policies for several low-dimensional tasks. For most manipulation tasks, we treat success rate as the primary performance metric, while Push-T is evaluated by target-area coverage. To measure inference efficiency, we report the FLOPs required for action generation alongside the speedup relative to the full DP-T model.

\medskip
\noindent\textbf{Baselines.}
We leverage the full DP-T policy as the full-precision baseline and subsequently compare EVO with two representative training-free caching methods. EfficientVLA refreshes features
at fixed denoising intervals as a static scheduling paradigm, while BAC achieves local adaptive scheduling through a block-wise update policy and an error-propagation repair mechanism. Distinct from prior works, EVO globally searches for cache schedules over the full block-timestep lattice with more adaptive and fast performance.

\begin{table}[t]
\centering
\caption{Benchmark on Mixed Human (MH) demonstration data. Success rates are evaluated by mean $\pm$ standard deviation over three evaluation seeds, and speedups are measured with full DP-T inference. EVO preserves the full-model average success rate of 0.79 while achieving up to $7.92\times$ speedup.}
\label{tab:mh_results}
\setlength{\tabcolsep}{4pt}
\resizebox{\textwidth}{!}{
\begin{tabular}{lccccccc}
\toprule
\multirow{2}{*}{Method}
& \multicolumn{4}{c}{Success Rate $\uparrow$}
& \multirow{2}{*}{AVG}
& \multirow{2}{*}{FLOPs}
& \multirow{2}{*}{Speed $\times$} \\
\cmidrule(lr){2-5}
& Lift & Can & Square & Transport & & & \\
\midrule
Full Precision
& $\mathbf{1.00} \pm 0.00$
& $0.94 \pm 0.02$
& $0.69 \pm 0.14$
& $\mathbf{0.51} \pm 0.01$
& \textbf{0.79} & 15.77G & -- \\
\midrule
EfficientVLA ($M=8$)
& $\mathbf{1.00} \pm 0.00$
& $0.72 \pm 0.09$
& $0.53 \pm 0.06$
& $0.00 \pm 0.00$
& 0.56 & 1.64G & 9.62 \\
EfficientVLA ($M=10$)
& $0.75 \pm 0.05$
& $0.09 \pm 0.04$
& $0.02 \pm 0.02$
& $0.00 \pm 0.00$
& 0.22 & 1.95G & 8.09 \\
\midrule
BAC ($M=8$)
& $\mathbf{1.00} \pm 0.00$
& $0.73 \pm 0.08$
& $0.61 \pm 0.06$
& $0.45 \pm 0.06$
& 0.70 & 2.32G & 6.80 \\
BAC ($M=10$)
& $\mathbf{1.00} \pm 0.00$
& $0.94 \pm 0.02$
& $0.75 \pm 0.06$
& $\mathbf{0.47} \pm 0.10$
& \textbf{0.79} & 2.71G & 5.82 \\
\midrule
\rowcolor{evoshade}
EVO ($M=8$)
& $\mathbf{1.00} \pm 0.00$
& $\mathbf{0.96} \pm 0.02$
& $0.71 \pm 0.06$
& $\mathbf{0.47} \pm 0.03$
& \textbf{0.79} & 1.99G & 7.92 \\
\rowcolor{evoshade}
EVO ($M=10$)
& $\mathbf{1.00} \pm 0.00$
& $0.93 \pm 0.03$
& $\mathbf{0.77} \pm 0.05$
& $\mathbf{0.47} \pm 0.03$
& \textbf{0.79} & 2.34G & 6.74 \\
\bottomrule
\end{tabular}
}
\end{table}

\medskip
\noindent\textbf{Implementation Details.}
All methods are evaluated by the final checkpoint of the publicly released
pretrained DP-T policy, with the denoising process set to 100 steps. EVO adopts a unified genetic-search configuration with population size, number of elites, and
maximum number of generations set to 30, 10, and 30, respectively. Unless
otherwise specified, we evaluate two cache-budget settings, $M=8$ and $M=10$,
where $M$ denotes the average number of refresh steps per cacheable module.
Since DP-T contains 24 cacheable modules, these settings correspond to total
budgets of $K=192$ and $K=240$ block-timestep update positions, respectively. During the 
search, each candidate schedule $\mathcal{S}$ is first evaluated by 20 rollout
episodes. Candidates satisfying the target trigger are then confirmed by 50
rollout episodes with independent random seeds. Search stops early when the
confirmed score satisfies the target performance criterion. Following the search phase, the selected cache schedule is frozen and deployed for the final inference evaluation. Final results are evaluated over three random seeds, with 50 test episodes per seed. All experiments run on a workstation equipped with an NVIDIA GeForce RTX 4090D GPU.

\begin{table}[H]
\centering
\caption{Benchmark on multi-stage manipulation tasks. For Block Push, $p_x$ denotes the frequency of pushing $x$ blocks into the targets. For Kitchen, $p_x$ denotes the frequency of interacting with $x$ or more objects.}
\label{tab:multi_stage_results}
\setlength{\tabcolsep}{4pt}
\resizebox{\textwidth}{!}{
\begin{tabular}{lccccccccc}
\toprule
\multirow{2}{*}{Method}
& \multicolumn{6}{c}{Success Rate $\uparrow$}
& \multirow{2}{*}{AVG}
& \multirow{2}{*}{FLOPs}
& \multirow{2}{*}{Speed $\times$} \\
\cmidrule(lr){2-7}
& BP$_{p_1}$ & BP$_{p_2}$ & Kit$_{p_1}$ & Kit$_{p_2}$ & Kit$_{p_3}$ & Kit$_{p_4}$ & & & \\
\midrule
Full Precision
& $0.99 \pm 0.01$ & $0.95 \pm 0.01$ & $\mathbf{1.00} \pm 0.00$ & $\mathbf{1.00} \pm 0.00$ & $\mathbf{1.00} \pm 0.00$ & $\mathbf{0.99} \pm 0.01$ & \textbf{0.99} & 15.77G & -- \\
\midrule
EfficientVLA ($M=8$)
& $0.99 \pm 0.01$ & $0.93 \pm 0.01$ & $0.18 \pm 0.04$ & $0.04 \pm 0.02$ & $0.02 \pm 0.00$ & $0.00 \pm 0.00$ & 0.36 & 1.64G & 9.62 \\
EfficientVLA ($M=10$)
& $0.98 \pm 0.02$ & $0.91 \pm 0.08$ & $0.29 \pm 0.02$ & $0.00 \pm 0.00$ & $0.00 \pm 0.00$ & $0.00 \pm 0.00$ & 0.36 & 1.95G & 8.09 \\
\midrule
BAC ($M=8$)
& $0.99 \pm 0.01$ & $0.93 \pm 0.02$ & $0.99 \pm 0.01$ & $0.99 \pm 0.01$ & $0.98 \pm 0.02$ & $0.90 \pm 0.02$ & 0.97 & 2.33G & 6.77 \\
BAC ($M=10$)
& $\mathbf{1.00} \pm 0.00$ & $\mathbf{0.97} \pm 0.03$ & $\mathbf{1.00} \pm 0.00$ & $\mathbf{1.00} \pm 0.00$ & $\mathbf{1.00} \pm 0.00$ & $0.96 \pm 0.04$ & \textbf{0.99} & 2.67G & 5.91 \\
\midrule
\rowcolor{evoshade}
EVO ($M=8$)
& $0.97 \pm 0.03$ & $0.95 \pm 0.06$ & $0.99 \pm 0.01$ & $0.98 \pm 0.02$ & $0.98 \pm 0.02$ & $0.93 \pm 0.03$ & 0.97 & 1.97G & 8.01 \\
\rowcolor{evoshade}
EVO ($M=10$)
& $0.99 \pm 0.01$ & $\mathbf{0.97} \pm 0.03$ & $\mathbf{1.00} \pm 0.00$ & $0.99 \pm 0.01$ & $0.99 \pm 0.01$ & $\mathbf{0.97} \pm 0.02$ & \textbf{0.99} & 2.24G & 7.06 \\
\bottomrule
\end{tabular}
}
\end{table}

\subsection{Main Results}

Tables~\ref{tab:ph_results}, \ref{tab:mh_results}, and \ref{tab:multi_stage_results} show that EVO substantially accelerates DP-T inference while preserving
closed-loop control performance close to the full model. We control the
computational budget by the number of cache update steps $M$. On the PH,
MH, and multi-stage tasks, EVO achieves peak average success rates of 0.78, 0.79,
and 0.99, respectively, close to the full DP-T results of 0.80, 0.79, and 0.99.
Meanwhile, EVO consistently achieves stable acceleration rates above 6.7× across most of the tasks, while reducing the FLOPs of action-generation from 15.77G to about
1.96G--2.34G. Wall-clock measurements on five representative tasks further show that EVO reduces the average policy inference time from 448.91 ms to 132.17 ms, increasing the corresponding inference frequency from 2.23 Hz to 7.57 Hz. Detailed results are provided in Sec.~A.7 of the supplementary material.

Compared with EfficientVLA, EVO substantially reduces performance collapse on
challenging tasks such as Transport, Tool, and Kitchen. Moreover, EVO achieves comparable or better average success rates of BAC with lower FLOPs, especially on PH and MH tasks, exhibiting a much stronger trade-off between task performance and inference efficiency.
These results indicate that rollout-driven global block-timestep search allocates the
fixed cache-update budget more effectively than fixed-interval refreshing or
local adaptive updating schedules. Representative refresh schedules for BAC and EVO are reported in Sec.~A.5 of the supplementary material.

\begin{table}[H]
\centering
\caption{Ablation study on EVO.}
\label{tab:ablation_evo}
\setlength{\tabcolsep}{4pt}
\resizebox{\textwidth}{!}{
\begin{tabular}{lccccccccc}
\toprule
\multirow{2}{*}{Method}
& \multicolumn{5}{c}{Success Rate $\uparrow$}
& \multirow{2}{*}{AVG}
& \multirow{2}{*}{Speed $\times$}
& \multirow{2}{*}{GPU-days}
& \multirow{2}{*}{Search Steps} \\
\cmidrule(lr){2-6}
& Lift$_{ph}$ & Can$_{ph}$ & Square$_{ph}$ & Lift$_{mh}$ & Can$_{mh}$ & & & & \\
\midrule
Block-wise GA
& $0.95 \pm 0.05$
& $0.90 \pm 0.02$
& $0.81 \pm 0.05$
& $0.95 \pm 0.02$
& $0.76 \pm 0.10$
& 0.87 & 8.17 & 1.48 & 358 \\

Block-Step GA
& $\mathbf{1.00} \pm 0.00$
& $0.95 \pm 0.01$
& $0.82 \pm 0.00$
& $0.97 \pm 0.01$
& $0.83 \pm 0.09$
& 0.91 & 8.05 & 1.00 & 242 \\

Block-wise GA + RI
& $0.99 \pm 0.01$
& $0.91 \pm 0.02$
& $0.86 \pm 0.04$
& $\mathbf{1.00} \pm 0.00$
& $0.89 \pm 0.05$
& 0.93 & 8.26 & 1.40 & 226 \\
\midrule
\rowcolor{evoshade}
Block-Step GA + RI
& $\mathbf{1.00} \pm 0.00$
& $\mathbf{0.98} \pm 0.04$
& $\mathbf{0.88} \pm 0.02$
& $\mathbf{1.00} \pm 0.00$
& $\mathbf{0.96} \pm 0.02$
& \textbf{0.96} & 8.01 & 0.96 & 222 \\
\bottomrule
\end{tabular}
}
\end{table}

\subsection{Ablation Study}
\label{sec:ablstudy}
To validate EVO's key design choices, we conduct ablation
studies on representative PH and MH tasks, with results in
\mbox{Table~\ref{tab:ablation_evo}}. Since EVO optimizes cache schedules by an offline genetic algorithm (GA), we report GPU-days~\cite{chen2019progressive} and Search Steps
as search-cost metrics in addition to final task performance and inference speed. Notably, Search Steps abbreviates Search Steps to Target and denotes the average
number of candidate schedules evaluated before reaching the target performance.

\noindent\textbf{Ablation Study Methods.}
We consider three variants of EVO to evaluate two key designs: global block--timestep budget allocation and redundancy-aware initialization (RI). All variants follow
the same genetic search procedure and differ only in the schedule-space
constraint and initialization strategy. To evaluate the effect of global block--timestep budget allocation, we define \textit{Block-wise GA}, which keeps the genetic search unchanged but requires
exactly eight refresh steps per cacheable block, thereby assigning the same budget to each block. We then define \textit{Block-Step GA}, which removes the constraint of per-block budget and allows the same total budget to be freely distributed over
the full block--timestep lattice. To evaluate the effectiveness of RI, we apply redundancy-aware initialization to the above two search spaces, yielding \textit{Block-wise GA + RI} and \textit{Block-Step GA + RI}. The latter corresponds to the full EVO design.

\noindent\textbf{Effectiveness of Global Block-timestep Allocation.}
To validate the role of global block--timestep budget reallocation, we compare
\textit{Block-wise GA} and \textit{Block-Step GA}. The results show that
\textit{Block-Step GA} improves the average success rate from 0.87 to 0.91, while
reducing the search cost from 1.48 to 1.00 GPU-days and decreasing Search Steps
from 358 to 242. This indicates that equal per-block budget limit schedule
quality, whereas full block--timestep allocation utilize the fixed update budget more effectively.

\noindent\textbf{Effectiveness of Redundancy-aware Initialization.}
To validate the role of RI, we add redundancy-aware initialization to both
search spaces. RI improves the average success rate of \textit{Block-wise GA} from 0.87
to 0.93 and that of \textit{Block-Step GA} from 0.91 to 0.96. The full EVO design, corresponding to \textit{Block-Step GA + RI}, achieves the highest average success
rate, the lowest search cost of 0.96 GPU-days, and the fewest Search Steps of
222. This shows that RI provides more effective initial candidates for genetic
search, and combining it with cross-block budget reallocation yields the best
overall performance. We further investigate the choice of feature dissimilarity metric and the role of random individuals during the  initialization procedure. Specificly, the methods with cosine dissimilarity achieve the highest average success rate of 0.79, while retaining random individuals slightly improves average performance and reduces the average number of search steps from 270 to 240. Detailed results are provided in Secs.~A.2 and A.3 of the supplementary material.

\noindent\textbf{Offline Search Cost and Amortization.}
As reported in Table~4, full EVO achieves the lowest offline search cost and the fewest Search Steps among all variants. The search is performed once per task, with all candidates evaluated offline in simulation using their mean performance over multiple environment seeds. For real-world applications, this procedure can be completed before deployment, and the selected schedule can then be deployed on the robot without additional online search or optimization overhead. Therefore, the selected schedule is robust to different initializations and can be reused across episodes without online search or adaptation. Its one-time cost is amortized over deployment because every action query benefits from reduced FLOPs and latency.

\section{Conclusion}
In this paper, we propose EVO, a novel training-free acceleration method for
transformer-based diffusion policy inference. EVO formulates cache scheduling as
a global resource allocation problem over the block--timestep lattice, and optimizes cache update positions through evolutionary search according to downstream
rollout performance. With redundancy-aware initialization and target-conditioned early stopping, EVO efficiently finds reliable offline cache schedules. Extensive experiments show that EVO substantially reduces the computation cost of action generation while
maintaining near-full DP-T control performance, achieving up to about
8$\times$ inference speedup. 

\section{Acknowledgements}
This work was supported in part by the Fundamental Research Funds for the Central Universities (No. XJSJ25005), the Outstanding Youth Science Foundation of Shaanxi Province under Grant 2025JC-JCQN-083, the Natural Science Foundation of Xi'an under Grant 2025JH-ZRKX-0540, and the Fundamental Research Funds for the Central Universities under Grant QTZX26145. Thanks to the help provided by the National Experimental Teaching Demonstration Center for Computer Network and Information Security affiliated with Xidian University.

\bibliographystyle{splncs04}
\bibliography{references}

@article{chi2025diffusion,
  title={Diffusion policy: Visuomotor policy learning via action diffusion},
  author={Chi, Cheng and Xu, Zhenjia and Feng, Siyuan and Cousineau, Eric and Du, Yilun and Burchfiel, Benjamin and Tedrake, Russ and Song, Shuran},
  journal={The International Journal of Robotics Research},
  volume={44},
  number={10-11},
  pages={1684--1704},
  year={2025},
  publisher={Sage Publications Sage UK: London, England}
}

@article{ho2020denoising,
  title={Denoising diffusion probabilistic models},
  author={Ho, Jonathan and Jain, Ajay and Abbeel, Pieter},
  journal={Advances in neural information processing systems},
  volume={33},
  pages={6840--6851},
  year={2020}
}

@INPROCEEDINGS{Ze-RSS-24, 
    AUTHOR    = {Yanjie Ze AND Gu Zhang AND Kangning Zhang AND Chenyuan Hu AND Muhan Wang AND Huazhe Xu}, 
    TITLE     = {{3D Diffusion Policy: Generalizable Visuomotor Policy Learning via Simple 3D Representations}}, 
    BOOKTITLE = {Proceedings of Robotics: Science and Systems}, 
    YEAR      = {2024}, 
    ADDRESS   = {Delft, Netherlands}, 
    MONTH     = {July}
}

@article{brohan2023rt,
  title={RT-1: Robotics Transformer for Real-World Control at Scale},
  author={Brohan, Anthony and Brown, Noah and Carbajal, Justice and Chebotar, Yevgen and Dabis, Joseph and Finn, Chelsea and Gopalakrishnan, Keerthana and Hausman, Karol and Herzog, Alexander and Hsu, Jasmine and others},
  journal={Robotics: Science and Systems XIX},
  year={2023},
  publisher={Robotics: Science and Systems Foundation}
}

@inproceedings{peebles2023scalable,
  title={Scalable diffusion models with transformers},
  author={Peebles, William and Xie, Saining},
  booktitle={Proceedings of the IEEE/CVF international conference on computer vision},
  pages={4195--4205},
  year={2023}
}

@inproceedings{ma2024deepcache,
  title={Deepcache: Accelerating diffusion models for free},
  author={Ma, Xinyin and Fang, Gongfan and Wang, Xinchao},
  booktitle={Proceedings of the IEEE/CVF conference on computer vision and pattern recognition},
  pages={15762--15772},
  year={2024}
}

@inproceedings{lyu2025fastercache,
  title={Fastercache: Training-free video diffusion model acceleration with high quality},
  author={Lyu, Zhengyao and Si, Chenyang and Song, Junhao and Yang, Zhenyu and Qiao, Yu and Liu, Ziwei and Wong, Kwan-Yee K},
  booktitle={International Conference on Learning Representations},
  volume={2025},
  pages={33132--33156},
  year={2025}
}

@article{
liu2025faster,
title={Faster Diffusion Through Temporal Attention Decomposition},
author={Haozhe Liu and Wentian Zhang and Jinheng Xie and Francesco Faccio and Mengmeng Xu and Tao Xiang and Mike Zheng Shou and Juan-Manuel Perez-Rua and J{\"u}rgen Schmidhuber},
journal={Transactions on Machine Learning Research},
issn={2835-8856},
year={2025},
note={}
}

@article{ma2024learning,
  title={Learning-to-cache: Accelerating diffusion transformer via layer caching},
  author={Ma, Xinyin and Fang, Gongfan and Bi Mi, Michael and Wang, Xinchao},
  journal={Advances in Neural Information Processing Systems},
  volume={37},
  pages={133282--133304},
  year={2024}
}

@inproceedings{zhao2025real,
  title={Real-time video generation with pyramid attention broadcast},
  author={Zhao, Xuanlei and Jin, Xiaolong and Wang, Kai and You, Yang},
  booktitle={International Conference on Learning Representations},
  volume={2025},
  pages={3296--3319},
  year={2025}
}

@inproceedings{
zou2025accelerating,
title={Accelerating Diffusion Transformers with Token-wise Feature Caching},
author={Chang Zou and Xuyang Liu and Ting Liu and Siteng Huang and Linfeng Zhang},
booktitle={The Thirteenth International Conference on Learning Representations},
year={2025}
}

@article{yang2026efficientvla,
  title={Efficientvla: Training-free acceleration and compression for vision-language-action models},
  author={Yang, Yantai and Wang, Yuhao and Wen, Zichen and Zhongwei, Luo and Zou, Chang and Zhang, Zhipeng and Wen, Chuan and Zhang, Linfeng},
  journal={Advances in Neural Information Processing Systems},
  volume={38},
  pages={40891--40914},
  year={2026}
}

@inproceedings{
ji2026blockwise,
title={Block-wise Adaptive Caching for Accelerating Diffusion Policy},
author={Kangye Ji and Yuan Meng and Hanyun Cui and Ye Li and Jianbo Zhou and Shengjia Hua and Lei Chen and Zhi Wang},
booktitle={The Fourteenth International Conference on Learning Representations},
year={2026}
}

@inproceedings{wang2025one,
  title={One-Step Diffusion Policy: Fast Visuomotor Policies via Diffusion Distillation},
  author={Wang, Zhendong and Li, Max and Mandlekar, Ajay and Xu, Zhenjia and Fan, Jiaojiao and Narang, Yashraj and Fan, Linxi and Zhu, Yuke and Balaji, Yogesh and Zhou, Mingyuan and others},
  booktitle={International Conference on Machine Learning},
  pages={63399--63416},
  year={2025},
  organization={PMLR}
}

@inproceedings{janner2022planning,
  title={Planning with Diffusion for Flexible Behavior Synthesis},
  author={Janner, Michael and Du, Yilun and Tenenbaum, Joshua and Levine, Sergey},
  booktitle={International Conference on Machine Learning},
  pages={9902--9915},
  year={2022},
  organization={PMLR}
}

@inproceedings{rombach2022high,
  title={High-resolution image synthesis with latent diffusion models},
  author={Rombach, Robin and Blattmann, Andreas and Lorenz, Dominik and Esser, Patrick and Ommer, Bj{\"o}rn},
  booktitle={Proceedings of the IEEE/CVF conference on computer vision and pattern recognition},
  pages={10684--10695},
  year={2022}
}

@InProceedings{pmlr-v164-mandlekar22a,
  title = {What Matters in Learning from Offline Human Demonstrations for Robot Manipulation},
  author = {Mandlekar, Ajay and Xu, Danfei and Wong, Josiah and Nasiriany, Soroush and Wang, Chen and Kulkarni, Rohun and Fei-Fei, Li and Savarese, Silvio and Zhu, Yuke and Mart\'in-Mart\'in, Roberto},
  booktitle = 	 {Proceedings of the 5th Conference on Robot Learning},
  pages = 	 {1678--1690},
  year = 	 {2022},
}

@inproceedings{
song2021denoising,
title={Denoising Diffusion Implicit Models},
author={Jiaming Song and Chenlin Meng and Stefano Ermon},
booktitle={International Conference on Learning Representations},
year={2021},
}

@article{lu2022dpm,
  title={Dpm-solver: A fast ode solver for diffusion probabilistic model sampling in around 10 steps},
  author={Lu, Cheng and Zhou, Yuhao and Bao, Fan and Chen, Jianfei and Li, Chongxuan and Zhu, Jun},
  journal={Advances in neural information processing systems},
  volume={35},
  pages={5775--5787},
  year={2022}
}

@article{lu2025dpm,
  title={Dpm-solver++: Fast solver for guided sampling of diffusion probabilistic models},
  author={Lu, Cheng and Zhou, Yuhao and Bao, Fan and Chen, Jianfei and Li, Chongxuan and Zhu, Jun},
  journal={Machine Intelligence Research},
  volume={22},
  number={4},
  pages={730--751},
  year={2025},
  publisher={Springer}
}

@article{zhao2023unipc,
  title={Unipc: A unified predictor-corrector framework for fast sampling of diffusion models},
  author={Zhao, Wenliang and Bai, Lujia and Rao, Yongming and Zhou, Jie and Lu, Jiwen},
  journal={Advances in Neural Information Processing Systems},
  volume={36},
  pages={49842--49869},
  year={2023}
}

@inproceedings{
  salimans2022progressive,
  title={Progressive Distillation for Fast Sampling of Diffusion Models},
  author={Tim Salimans and Jonathan Ho},
  booktitle={International Conference on Learning Representations},
  year={2022},
}

@article{ji2026sparse,
  title={Sparse ActionGen: Accelerating Diffusion Policy with Real-time Pruning},
  author={Ji, Kangye and Meng, Yuan and Jianbo, Zhou and Li, Ye and Cui, Hanyun and Wang, Zhi},
  journal={arXiv preprint arXiv:2601.12894},
  year={2026}
}

@InProceedings{pmlr-v202-song23a,
  title = 	 {Consistency Models},
  author =       {Song, Yang and Dhariwal, Prafulla and Chen, Mark and Sutskever, Ilya},
  booktitle = 	 {Proceedings of the 40th International Conference on Machine Learning},
  pages = 	 {32211--32252},
  year = 	 {2023},
  volume = 	 {202},
  series = 	 {Proceedings of Machine Learning Research},
  month = 	 {23--29 Jul},
  publisher =    {PMLR},
}

@inproceedings{wimbauer2024cache,
  title={Cache me if you can: Accelerating diffusion models through block caching},
  author={Wimbauer, Felix and Wu, Bichen and Schoenfeld, Edgar and Dai, Xiaoliang and Hou, Ji and He, Zijian and Sanakoyeu, Artsiom and Zhang, Peizhao and Tsai, Sam and Kohler, Jonas and others},
  booktitle={Proceedings of the IEEE/CVF Conference on Computer Vision and Pattern Recognition},
  pages={6211--6220},
  year={2024}
}

@article{selvaraju2024fora,
  title={Fora: Fast-forward caching in diffusion transformer acceleration},
  author={Selvaraju, Pratheba and Ding, Tianyu and Chen, Tianyi and Zharkov, Ilya and Liang, Luming},
  journal={arXiv preprint arXiv:2407.01425},
  year={2024}
}

@inproceedings{son2026relational,
  title={Relational Feature Caching for Accelerating Diffusion Transformers},
  author={Son, Byunggwan and Jeon, Jeimin and Choi, Jeongwoo and Ham, Bumsub},
  booktitle={The Fourteenth International Conference on Learning Representations},
  year={2026}
}

@article{zames1981genetic,
  title={Genetic algorithms in search, optimization and machine learning},
  author={Zames, G},
  journal={Inf Tech J},
  volume={3},
  number={1},
  pages={301},
  year={1981}
}

@inproceedings{real2019regularized,
  title={Regularized Evolution for Image Classifier Architecture Search},
  author={Real, Esteban and Aggarwal, Alok and Huang, Yanping and Le, Quoc V},
  booktitle={Proceedings of the AAAI Conference on Artificial Intelligence},
  volume={33},
  pages={4780--4789},
  year={2019}
}

@book{holland1992adaptation,
  title={Adaptation in natural and artificial systems: an introductory analysis with applications to biology, control, and artificial intelligence},
  author={Holland, John H},
  year={1992},
  publisher={MIT Press}
}

@article{stanley2002evolving,
  title={Evolving neural networks through augmenting topologies},
  author={Stanley, Kenneth O and Miikkulainen, Risto},
  journal={Evolutionary computation},
  volume={10},
  number={2},
  pages={99--127},
  year={2002},
  publisher={MIT Press}
}

@article{silva2015odneat,
  title={odNEAT: An algorithm for decentralised online evolution of robotic controllers},
  author={Silva, Fernando and Urbano, Paulo and Correia, Lu{\'\i}s and Christensen, Anders Lyhne},
  journal={Evolutionary Computation},
  volume={23},
  number={3},
  pages={421--449},
  year={2015},
  publisher={MIT Press}
}

@inproceedings{guo2025neural,
  title={Neural Architecture Search-Based Meta-Reinforcement Learning in Robotic Simulation Environments},
  author={Guo, Jun and Zheng, Xiao and Chen, Yuanfang and Zhang, Xianchao},
  booktitle={2025 International Joint Conference on Neural Networks (IJCNN)},
  pages={1--10},
  year={2025},
  organization={IEEE}
}

@inproceedings{hegde2023efficiently,
  title={Efficiently learning small policies for locomotion and manipulation},
  author={Hegde, Shashank and Sukhatme, Gaurav S},
  booktitle={2023 IEEE International Conference on Robotics and Automation (ICRA)},
  pages={5909--5915},
  year={2023},
  organization={IEEE}
}

@inproceedings{chen2019progressive,
  title={Progressive differentiable architecture search: Bridging the depth gap between search and evaluation},
  author={Chen, Xin and Xie, Lingxi and Wu, Jun and Tian, Qi},
  booktitle={Proceedings of the IEEE/CVF international conference on computer vision},
  pages={1294--1303},
  year={2019}
}

\end{document}


\appendix

\section{Supplemental Material}

In this supplemental material, we provide additional experimental details and analyses to support the main conclusions of the paper. In Appendix~A.1, we introduce the benchmark tasks and pretrained checkpoints used in our experiments, including RoboMimic, Push-T, Block Push, and Kitchen. In Appendix~A.2, we present additional experiments on the choice of similarity metrics for redundancy-aware initialization, analyzing how different feature dissimilarity measures affect the quality of the initial population and the subsequent search process. Appendix~A.3 further investigates the role of random individuals in the initialization stage, discussing the balance between guided sampling and population diversity.

We then provide additional visual analyses to further support the motivation of rollout-driven global cache scheduling. Appendix~A.4 presents single-block cache stress tests and the relationship between feature similarity and task sensitivity on representative tasks. These results show that task sensitivity is non-uniform across transformer blocks and module types, and that feature similarity, although useful as an initialization prior, is insufficient as the final scheduling objective. Finally, Appendix~A.5 compares the update locations obtained by BAC and EVO under comparable cache budgets, illustrating how EVO distributes refresh positions over the block--timestep lattice through rollout-driven global search.

\subsection{Additional Benchmark Details}

\subsubsection{Datasets}
\mbox{}\\[-1.2em]

\noindent\textbf{Push-T.}
Push-T is adapted from the implicit behavioral cloning benchmark and is used as a contact-rich planar pushing benchmark in Diffusion Policy. The dataset contains human demonstrations for pushing a T-shaped block under visual observations. Following the standard protocol, we evaluate Push-T by target-area coverage rather than binary success rate.

\noindent\textbf{Block Push.}
Block Push is adapted from the Behavior Transformer and IBC benchmarks. The dataset contains scripted demonstrations for pushing two blocks into two target regions. Since the two blocks can be pushed in different valid orders, this benchmark is commonly used to evaluate multi-modal action prediction.

\noindent\textbf{Kitchen.}
Kitchen is adapted from the Relay Policy Learning benchmark. It contains multi-step manipulation demonstrations in a simulated kitchen environment, where the policy interacts with multiple objects within a long-horizon rollout. We report \(p_x\), which denotes the frequency of completing at least \(x\) subtasks.

\noindent\textbf{RoboMimic.}
RoboMimic is a robot learning benchmark built from human teleoperation demonstrations. We evaluate EVO on five image-based tasks: Lift, Can, Square, Transport, and Tool-Hang. The benchmark provides Proficient Human (PH) demonstrations for these tasks and Mixed Human (MH) demonstrations for four of them. Compared with PH demonstrations, MH demonstrations contain both proficient and non-proficient human demonstrations, leading to more diverse data distributions.

\subsubsection{Pre-trained Checkpoints and Evaluation Protocol}
\mbox{}\\[-1.2em]

We use the publicly released \texttt{diffusion\_policy\_transformer} checkpoints provided by Diffusion Policy. All experiments are conducted in simulation using the final checkpoint of each pre-trained policy. We do not train or fine-tune the original policies. EVO only modifies the inference-time cache schedule, while keeping the policy weights and the denoising sampler unchanged.

For the offline search stage, each task is searched with one search seed. Candidate schedules are first evaluated using a lightweight quick evaluation, and schedules satisfying the target-trigger condition are further verified by a confirmation evaluation using an independent seed. After search, the selected cache schedule is fixed for final evaluation. For the reported results, we use three testing seeds and run 50 test episodes for each seed. We report the mean performance and the standard deviation across the three testing seeds.

\subsection{Additional Experiments on Different Similarity Metrics}

In this experiment, we evaluate how the choice of feature dissimilarity metric affects the redundancy-aware initialization in EVO. Specifically, we consider four commonly used metrics: Mean Squared Error (MSE), \(L_1\) distance, Wasserstein-1 distance (Wa), and Cosine dissimilarity (Cosine). These metrics are used only to construct the dissimilarity scores for initializing the population, while the subsequent genetic search configuration and cache budget are kept unchanged for a fair comparison.

Due to the computational cost of offline search, we conduct this analysis on four representative tasks covering both Proficient Human (PH) and Mixed Human (MH) demonstrations. Table~\ref{tab:metric_ablation} reports the success rate as mean \(\pm\) standard deviation over three evaluation seeds, together with the average success rate, inference FLOPs, and Search Steps. 

The results show that the similarity metric used for initialization can affect both the quality of the searched schedule and the efficiency of the search process. A more suitable metric provides higher-quality initial candidates, allowing the evolutionary search to reach effective schedules with fewer evaluated candidates while preserving downstream task performance. Since all variants use the same cache budget, their inference FLOPs remain comparable, and the main differences come from the final searched schedules and the search cost.

\begin{table}[H]
\centering
\caption{Benchmark results on four representative tasks using different similarity metrics for similarity-guided initialization.}
\label{tab:metric_ablation}
\setlength{\tabcolsep}{4pt}
\resizebox{\textwidth}{!}{
\begin{tabular}{lccccccc}
\toprule
\multirow{2}{*}{Method}
& \multicolumn{4}{c}{Success Rate $\uparrow$}
& \multirow{2}{*}{AVG}
& \multirow{2}{*}{FLOPs}
& \multirow{2}{*}{Search Steps} \\
\cmidrule(lr){2-5}
& Lift$_{ph}$ & Can$_{ph}$ & Square$_{mh}$ & Transport$_{mh}$ & & & \\
\midrule
EVO (MSE)
& $0.99 \pm 0.01$ & $0.95 \pm 0.03$ & $0.68 \pm 0.02$ & $0.37 \pm 0.10$ & 0.75 & 1.96G & 240 \\
EVO ($L_1$)
& $0.99 \pm 0.01$ & $0.94 \pm 0.02$ & $0.45 \pm 0.07$ & $0.35 \pm 0.07$ & 0.68 & 1.96G & 363 \\
EVO (Wa)
& $0.99 \pm 0.01$ & $0.95 \pm 0.03$ & $0.53 \pm 0.02$ & $0.41 \pm 0.04$ & 0.72 & 1.98G & 463 \\
EVO (Cosine)
& $1.00 \pm 0.00$ & $0.98 \pm 0.04$ & $0.71 \pm 0.06$ & $0.47 \pm 0.03$ & 0.79 & 2.05G & 240 \\
\bottomrule
\end{tabular}
}
\end{table}

\subsection{Effect of Random Individuals in Initialization}

We further evaluate the effect of random individuals in the redundancy-aware initialization of EVO. The purpose of this experiment is to examine whether guided sampling alone is sufficient for constructing the initial population. We compare the full initialization strategy with a variant that removes random individuals, while keeping the cache budget and genetic search configuration unchanged.

Table~\ref{tab:random_individuals} reports the results on four representative PH and MH tasks. We report success rates as mean \(\pm\) standard deviation over three evaluation seeds, together with AVG, FLOPs, and Search Steps. The comparison shows that random individuals help preserve population diversity and improve the robustness of the search process. This suggests that redundancy-aware initialization benefits from both biasing the
population toward less redundant block--timestep positions and retaining random
individuals for exploration.

\begin{table}[H]
\centering
\caption{Effect of random individuals in redundancy-aware initialization on four representative tasks.}
\label{tab:random_individuals}
\setlength{\tabcolsep}{4pt}
\resizebox{\textwidth}{!}{
\begin{tabular}{lccccccc}
\toprule
\multirow{2}{*}{Method}
& \multicolumn{4}{c}{Success Rate $\uparrow$}
& \multirow{2}{*}{AVG}
& \multirow{2}{*}{FLOPs}
& \multirow{2}{*}{Search Steps} \\
\cmidrule(lr){2-5}
& Lift$_{ph}$ & Can$_{ph}$ & Square$_{mh}$ & Transport$_{mh}$ & & & \\
\midrule
EVO w/o Random
& $1.00 \pm 0.00$ & $0.99 \pm 0.01$ & $0.71 \pm 0.04$ & $0.41 \pm 0.09$ & 0.78 & 2.02 & 270 \\
EVO w/ Random
& $1.00 \pm 0.00$ & $0.98 \pm 0.04$ & $0.71 \pm 0.06$ & $0.47 \pm 0.03$ & 0.79 & 2.05G & 240 \\
\bottomrule
\end{tabular}
}
\end{table}

\subsection{Additional Motivation Analysis}

We provide additional visualizations on five representative tasks to support the motivation of rollout-driven global cache scheduling. For each task, we include a single-block cache stress test and a comparison between adjacent-step feature similarity and task sensitivity. The stress-test results show that applying the same sparse update pattern to different transformer blocks leads to different success rate drops, indicating non-uniform task sensitivity across layers and module types. The similarity-sensitivity plots further show that adjacent-step feature similarity is only weakly correlated with downstream performance degradation. These observations suggest that feature similarity can serve as a useful prior for initialization, but is not sufficient as the final scheduling objective. This motivates EVO to combine redundancy-aware initialization with rollout-performance-driven global search.
\begin{figure}[H]
    \centering
    \begin{minipage}[t]{0.48\linewidth}
        \centering
        \resizebox{0.95\linewidth}{3.2cm}{
            \includegraphics{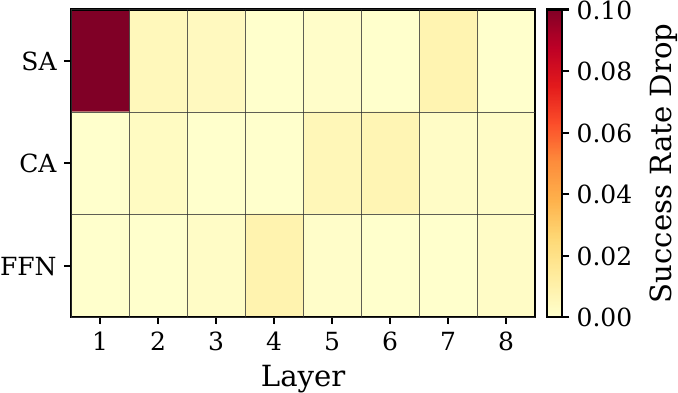}
        }
        \vspace{-0.4em}

        \small (a) Single-block cache stress test
    \end{minipage}
    \hfill
    \begin{minipage}[t]{0.48\linewidth}
        \centering
        \resizebox{0.95\linewidth}{3.2cm}{
            \includegraphics{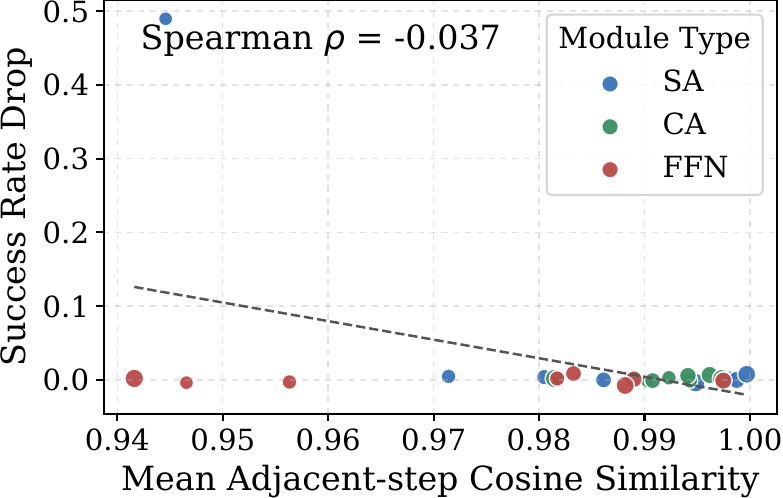}
        }
        \vspace{-0.4em}

        \small (b) Feature similarity vs. task sensitivity
    \end{minipage}
    \caption{Additional motivation analysis on Kitchen.}
    \label{fig:appendix_kitchen_motivation}
\end{figure}

\begin{figure}[H]
    \centering
    \begin{minipage}[t]{0.48\linewidth}
        \centering
        \resizebox{0.95\linewidth}{3.2cm}{
            \includegraphics{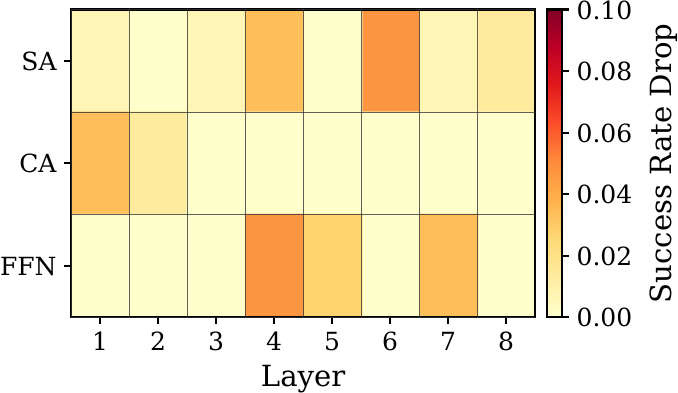}
        }
        \vspace{-0.4em}

        \small (a) Single-block cache stress test
    \end{minipage}
    \hfill
    \begin{minipage}[t]{0.48\linewidth}
        \centering
        \resizebox{0.95\linewidth}{3.2cm}{
            \includegraphics{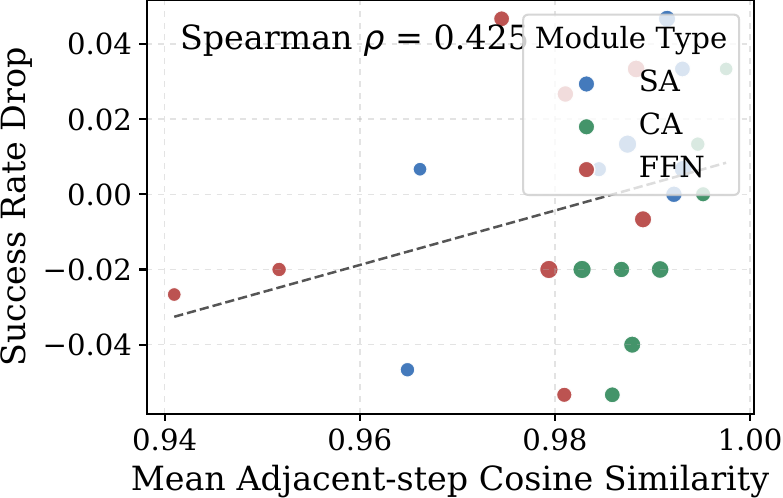}
        }
        \vspace{-0.4em}

        \small (b) Feature similarity vs. task sensitivity
    \end{minipage}
    \caption{Additional motivation analysis on Square MH.}
    \label{fig:appendix_square_mh_motivation}
\end{figure}

\begin{figure}[H]
    \centering
    \begin{minipage}[t]{0.48\linewidth}
        \centering
        \resizebox{0.95\linewidth}{3.2cm}{
            \includegraphics{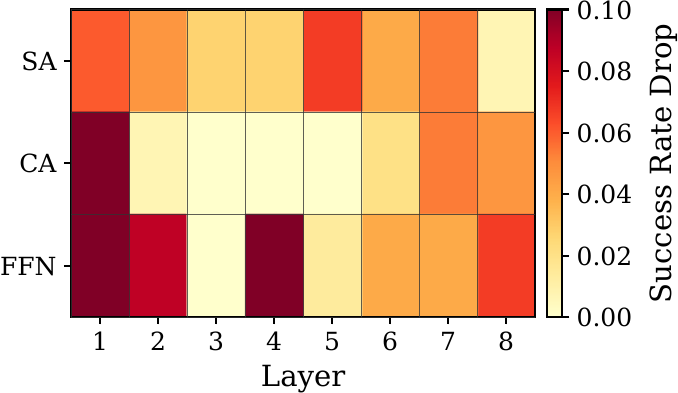}
        }
        \vspace{-0.4em}

        \small (a) Single-block cache stress test
    \end{minipage}
    \hfill
    \begin{minipage}[t]{0.48\linewidth}
        \centering
        \resizebox{0.95\linewidth}{3.2cm}{
            \includegraphics{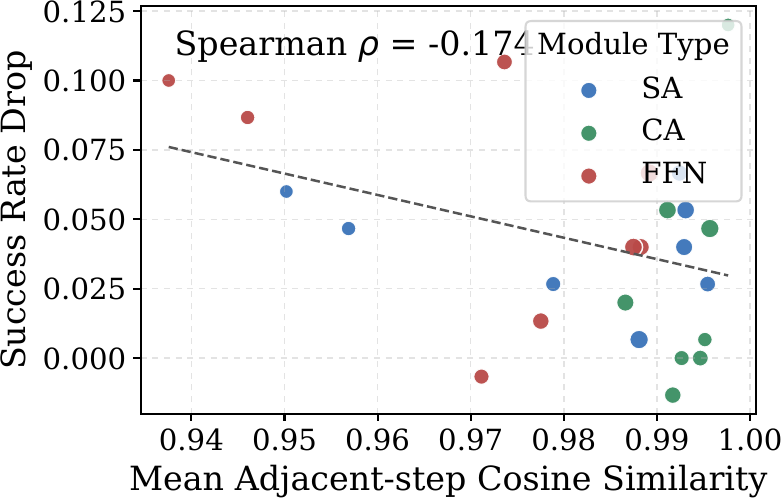}
        }
        \vspace{-0.4em}

        \small (b) Feature similarity vs. task sensitivity
    \end{minipage}
    \caption{Additional motivation analysis on Tool PH.}
    \label{fig:appendix_tool_hang_ph_motivation}
\end{figure}

\begin{figure}[H]
    \centering
    \begin{minipage}[t]{0.48\linewidth}
        \centering
        \resizebox{0.95\linewidth}{3.2cm}{
            \includegraphics{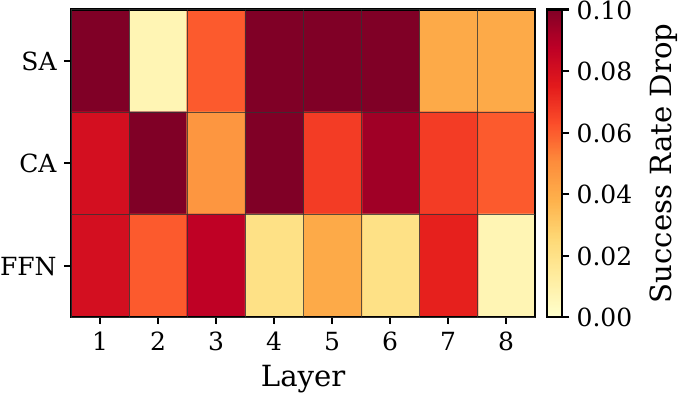}
        }
        \vspace{-0.4em}

        \small (a) Single-block cache stress test
    \end{minipage}
    \hfill
    \begin{minipage}[t]{0.48\linewidth}
        \centering
        \resizebox{0.95\linewidth}{3.2cm}{
            \includegraphics{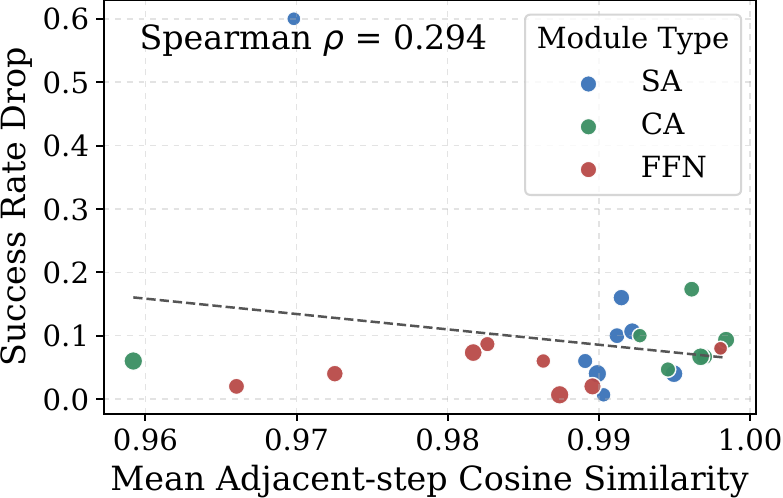}
        }
        \vspace{-0.4em}

        \small (b) Feature similarity vs. task sensitivity
    \end{minipage}
    \caption{Additional motivation analysis on Transport MH.}
    \label{fig:appendix_transport_mh_motivation}
\end{figure}

\begin{figure}[H]
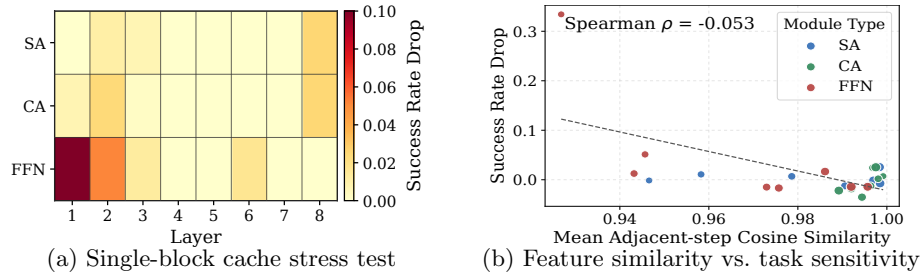

    \centering
    \begin{minipage}[t]{0.48\linewidth}
        \centering
        \resizebox{0.95\linewidth}{3.2cm}{
            \includegraphics{images/figure1a_single_block_stress.pdf}
        }
        \vspace{-0.4em}

        \small (a) Single-block cache stress test
    \end{minipage}
    \hfill
    \begin{minipage}[t]{0.48\linewidth}
        \centering
        \resizebox{0.95\linewidth}{3.2cm}{
            \includegraphics{images/figure1b_similarity_vs_sensitivity.pdf}
        }
        \vspace{-0.4em}

        \small (b) Feature similarity vs. task sensitivity
    \end{minipage}
    \caption{Additional motivation analysis on Push-T.}
    \label{fig:appendix_pusht_motivation}
\end{figure}

\subsection{Comparison of BAC and EVO Refresh Schedules}

Both BAC and EVO determine fixed update schedules offline and use the resulting
schedules during inference. To provide concrete details of the schedules used in our experiments, we report the update steps obtained by BAC and EVO on representative tasks under comparable cache budgets.

In Tables~\ref{tab:schedule_pusht}, \ref{tab:schedule_kitchen}, and
\ref{tab:schedule_lift_ph}, each row corresponds to a cacheable transformer block, and the two columns under \textit{Steps} list the update timesteps selected by BAC and EVO, respectively. This side-by-side comparison shows that BAC follows a block-wise adaptive update pattern, while EVO searches refresh positions over the full block--timestep lattice and therefore produces more flexible schedules across layers, module types, and denoising steps.

\setlength{\tabcolsep}{4pt}
\renewcommand{\arraystretch}{1.12}

\begin{longtable}{L{0.17\textwidth}L{0.36\textwidth}@{\hspace{1.2em}}L{0.36\textwidth}}
\caption{Update schedules of BAC and EVO on Push-T.}
\label{tab:schedule_pusht}\\

\toprule
\multirow{2}{*}[-0.35em]{\large\textbf{Block}}
& \multicolumn{2}{c}{\textbf{Steps}} \\
\cmidrule(lr){2-3}
& \multicolumn{1}{c}{\textbf{BAC}}
& \multicolumn{1}{c}{\textbf{EVO}} \\
\midrule
\endfirsthead

\toprule
\multirow{2}{*}[-0.35em]{\large\textbf{Block}}
& \multicolumn{2}{c}{\textbf{Steps}} \\
\cmidrule(lr){2-3}
& \multicolumn{1}{c}{\textbf{BAC}}
& \multicolumn{1}{c}{\textbf{EVO}} \\
\midrule
\endhead

\bottomrule
\endfoot

layers.0.SA
& 0, 4, 8, 14, 25, 32, 42, 52, 64, 73
& 12, 22, 24, 30, 38, 40, 42, 46, 49, 51, 57, 68, 73, 74, 77, 80, 83, 84, 86, 87 \\

layers.0.CA
& 0, 28, 49, 54, 65, 73, 79, 85, 91, 96
& 12, 18, 31, 32, 70, 76, 88, 99 \\

layers.0.FFN
& 0, 1, 4, 5, 7, 8, 14, 18, 26, 30, 31, 33, 34, 42, 51, 63, 64, 71, 76, 80, 81, 82, 86, 87, 88, 90, 91, 92, 94, 95, 97, 98
& 10, 14, 15, 16, 19, 20, 21, 25, 26, 30, 45, 64, 70, 72, 81, 83, 85, 89, 97 \\

layers.1.SA
& 0, 4, 15, 25, 32, 41, 52, 71, 81, 92
& 11, 26, 32, 35, 50, 65, 81, 82, 93 \\

layers.1.CA
& 0, 20, 32, 40, 46, 52, 60, 69, 86, 92
& 0, 11, 16, 17, 19, 21, 42, 44, 49, 60, 74, 83, 88, 90 \\

layers.1.FFN
& 0, 5, 14, 26, 34, 42, 51, 71, 81, 92
& 1, 2, 3, 8, 12, 13, 15, 21, 22, 32, 38, 51, 56, 58, 80, 83, 89, 91, 95 \\

layers.2.SA
& 0, 2, 13, 26, 41, 53, 63, 73, 81, 93
& 12, 41, 63, 69, 87 \\

layers.2.CA
& 0, 3, 23, 30, 41, 47, 56, 63, 73, 92
& 0, 6, 17, 81, 90, 96, 99 \\

layers.2.FFN
& 0, 1, 7, 31, 51, 63, 71, 80, 86, 92
& 1, 12, 18, 21, 28, 34, 37, 39, 43, 53, 57, 59, 78, 90, 91, 93, 95 \\

layers.3.SA
& 0, 9, 30, 42, 53, 66, 75, 83, 92, 97
& 19, 69, 95, 97, 99 \\

layers.3.CA
& 0, 10, 24, 29, 39, 46, 61, 73, 83, 92
& 1, 2, 4, 8, 12, 14, 29, 82, 84, 90, 95 \\

layers.3.FFN
& 0, 7, 33, 51, 71, 76, 81, 86, 92, 97
& 9, 10, 16, 19, 34, 41, 43, 45, 51, 54, 60, 62, 63, 76, 77 \\

layers.4.SA
& 0, 8, 41, 52, 71, 83, 90, 93, 96, 98
& 16, 40, 95, 98, 99 \\

layers.4.CA
& 0, 16, 24, 32, 42, 50, 63, 71, 85, 93
& 9, 10, 18, 47, 74, 83, 96, 98 \\

layers.4.FFN
& 0, 1, 7, 18, 26, 30, 51, 63, 64, 71, 76, 81, 82, 87, 88, 90, 91, 92, 94, 95, 97, 98
& 1, 4, 8, 21, 48, 49, 76, 82, 86, 95, 98 \\

layers.5.SA
& 0, 28, 62, 73, 83, 89, 92, 94, 96, 98
& 25, 31, 50, 99 \\

layers.5.CA
& 0, 14, 26, 41, 49, 62, 71, 83, 90, 96
& 16, 17, 37, 93, 95, 96 \\

layers.5.FFN
& 0, 1, 7, 18, 26, 30, 63, 64, 71, 76, 81, 82, 87, 88, 90, 92, 94, 95, 97, 98
& 3, 8, 17, 21, 35, 37, 38, 46, 53, 91, 93, 95, 98 \\

layers.6.SA
& 0, 29, 62, 76, 85, 90, 93, 95, 97, 98
& 29, 69, 97, 98, 99 \\

layers.6.CA
& 0, 9, 18, 26, 38, 44, 54, 63, 77, 90
& 14, 40, 66, 69, 90 \\

layers.6.FFN
& 0, 1, 7, 18, 30, 63, 81, 82, 88, 90, 92, 94, 95, 97
& 2, 5, 11, 18, 20, 27, 70, 78, 91, 95, 97, 99 \\

layers.7.SA
& 0, 10, 52, 71, 83, 89, 92, 94, 96, 98
& 19, 24, 91, 97, 98, 99 \\

layers.7.CA
& 0, 11, 33, 42, 53, 63, 73, 86, 93, 97
& 16, 96, 97, 98, 99 \\

layers.7.FFN
& 0, 1, 7, 18, 30, 81, 88, 92, 95, 97
& 0, 10, 13, 26, 67, 68, 77, 87, 95, 98, 99 \\

\end{longtable}

\setlength{\tabcolsep}{4pt}
\renewcommand{\arraystretch}{1.12}

\begin{longtable}{L{0.17\textwidth}L{0.36\textwidth}@{\hspace{1.2em}}L{0.36\textwidth}}
\caption{Update schedules of BAC and EVO on Kitchen.}
\label{tab:schedule_kitchen}\\

\toprule
\multirow{2}{*}[-0.35em]{\large\textbf{Block}}
& \multicolumn{2}{c}{\textbf{Steps}} \\
\cmidrule(lr){2-3}
& \multicolumn{1}{c}{\textbf{BAC}}
& \multicolumn{1}{c}{\textbf{EVO}} \\
\midrule
\endfirsthead

\toprule
\multirow{2}{*}[-0.35em]{\large\textbf{Block}}
& \multicolumn{2}{c}{\textbf{Steps}} \\
\cmidrule(lr){2-3}
& \multicolumn{1}{c}{\textbf{BAC}}
& \multicolumn{1}{c}{\textbf{EVO}} \\
\midrule
\endhead

\bottomrule
\endfoot

layers.0.SA
& 0, 1, 2, 3, 5, 6, 7, 8, 9, 10, 11, 13, 14, 18, 19, 21, 25, 27, 29, 30, 38, 41, 42, 43, 44, 46, 47, 54, 59, 64, 68, 70, 78, 79, 80, 83, 84, 86, 87, 89, 91, 92, 95, 97, 98
& 0, 3, 5, 9, 13, 19, 23, 24, 28, 29, 33, 40, 41, 47, 56, 57, 64, 68, 70, 71, 75, 78, 87, 92, 95, 97 \\

layers.0.CA
& 0, 1, 6, 18, 29, 41, 54, 68, 84, 94
& 16, 17, 25, 57, 59, 65, 68, 90, 98 \\

layers.0.FFN
& 0, 1, 2, 3, 5, 6, 7, 8, 9, 10, 11, 13, 14, 18, 21, 25, 27, 29, 30, 38, 41, 42, 44, 46, 47, 54, 59, 64, 68, 70, 78, 79, 80, 84, 86, 87, 89, 91, 92, 95, 97, 98
& 2, 4, 21, 44, 47, 60, 67, 73, 76, 78, 87, 91 \\

layers.1.SA
& 0, 4, 10, 19, 29, 42, 53, 62, 75, 87
& 3, 5, 13, 15, 20, 29, 49, 64, 94, 97 \\

layers.1.CA
& 0, 1, 4, 8, 11, 19, 30, 41, 55, 71
& 21, 28, 36, 72, 82 \\

layers.1.FFN
& 0, 5, 10, 21, 30, 42, 54, 68, 78, 87
& 1, 11, 21, 23, 24, 25, 36, 39, 48, 52, 56, 60, 64, 80, 85 \\

layers.2.SA
& 0, 4, 10, 19, 30, 41, 54, 72, 84, 92
& 4, 5, 7, 20, 21, 23, 81, 87, 90 \\

layers.2.CA
& 0, 2, 7, 13, 19, 30, 41, 54, 70, 87
& 2, 5, 11, 32, 34, 76, 89, 93, 98 \\

layers.2.FFN
& 0, 1, 5, 10, 18, 29, 47, 59, 78, 91
& 1, 3, 5, 16, 21, 24, 40, 45, 61, 74, 85, 86, 94 \\

layers.3.SA
& 0, 1, 7, 13, 19, 29, 41, 59, 82, 93
& 0, 2, 3, 5, 11, 18, 40, 57, 76, 86, 93, 94, 96 \\

layers.3.CA
& 0, 3, 5, 9, 13, 19, 33, 46, 68, 92
& 6, 14, 16, 19, 23, 54, 74, 79, 92, 94, 96 \\

layers.3.FFN
& 0, 2, 5, 8, 11, 18, 27, 41, 59, 78
& 0, 1, 9, 37, 40, 50, 61, 62, 89, 90 \\

layers.4.SA
& 0, 3, 8, 15, 25, 40, 59, 70, 81, 92
& 9, 14, 15, 17, 21, 26, 73, 76, 81, 99 \\

layers.4.CA
& 0, 3, 10, 18, 31, 42, 53, 62, 70, 82
& 1, 9, 32, 41, 49, 81, 92, 93 \\

layers.4.FFN
& 0, 2, 5, 9, 13, 18, 27, 41, 59, 79
& 1, 3, 17, 31, 33, 37, 43, 90, 95 \\

layers.5.SA
& 0, 3, 8, 15, 25, 37, 49, 62, 79, 91
& 1, 18, 30, 73, 74, 89, 91 \\

layers.5.CA
& 0, 2, 8, 17, 23, 32, 42, 53, 62, 72
& 3, 4, 7, 11, 13, 37, 85, 90 \\

layers.5.FFN
& 0, 1, 5, 6, 11, 18, 25, 27, 29, 38, 44, 47, 59, 64, 68, 80, 86, 89, 92, 95, 97, 98
& 3, 4, 7, 11, 16, 20, 22, 26, 46, 63, 74, 79, 93 \\

layers.6.SA
& 0, 8, 24, 36, 48, 58, 70, 84, 92, 98
& 16, 20, 24, 30, 76, 92 \\

layers.6.CA
& 0, 7, 17, 21, 29, 35, 41, 57, 80, 93
& 2, 3, 6, 32, 74 \\

layers.6.FFN
& 0, 1, 5, 6, 11, 25, 29, 38, 47, 59, 68, 86, 89, 95, 97, 98
& 0, 5, 8, 30, 67, 83, 93, 94, 98 \\

layers.7.SA
& 0, 1, 2, 6, 11, 25, 27, 38, 41, 59, 68, 75, 89, 90, 94, 95, 96, 98
& 41, 73, 97, 99 \\

layers.7.CA
& 0, 1, 7, 16, 29, 51, 60, 76, 88, 97
& 3, 10, 28, 31, 63, 74 \\

layers.7.FFN
& 0, 1, 6, 11, 25, 38, 59, 68, 89, 95
& 9, 15, 18, 20, 35, 45, 54, 68, 74, 80, 83, 93, 96 \\

\end{longtable}

\setlength{\tabcolsep}{4pt}
\renewcommand{\arraystretch}{1.12}

\begin{longtable}{L{0.17\textwidth}L{0.36\textwidth}@{\hspace{1.2em}}L{0.36\textwidth}}
\caption{Update schedules of BAC and EVO on Lift$_{ph}$.}
\label{tab:schedule_lift_ph}\\

\toprule
\multirow{2}{*}[-0.35em]{\large\textbf{Block}}
& \multicolumn{2}{c}{\textbf{Steps}} \\
\cmidrule(lr){2-3}
& \multicolumn{1}{c}{\textbf{BAC}}
& \multicolumn{1}{c}{\textbf{EVO}} \\
\midrule
\endfirsthead

\toprule
\multirow{2}{*}[-0.35em]{\large\textbf{Block}}
& \multicolumn{2}{c}{\textbf{Steps}} \\
\cmidrule(lr){2-3}
& \multicolumn{1}{c}{\textbf{BAC}}
& \multicolumn{1}{c}{\textbf{EVO}} \\
\midrule
\endhead

\bottomrule
\endfoot

layers.0.SA
& 0, 5, 14, 26, 43, 61, 80, 87, 93, 96
& 16, 92, 97 \\

layers.0.CA
& 0, 18, 28, 40, 53, 68, 80, 88, 94, 98
& 0, 8, 9, 19, 38, 48, 50, 51, 63, 91, 96 \\

layers.0.FFN
& 0, 3, 4, 5, 7, 8, 10, 11, 14, 16, 19, 21, 28, 29, 36, 42, 43, 46, 53, 56, 59, 62, 64, 66, 68, 69, 70, 71, 73, 75, 79, 80, 81, 83, 84, 86, 88, 90, 91, 92, 94, 96, 97, 98, 99
& 12, 14, 22, 44, 47, 54, 61, 77, 82, 95 \\

layers.1.SA
& 0, 3, 4, 5, 7, 8, 10, 11, 14, 16, 19, 26, 28, 29, 36, 42, 43, 46, 53, 56, 59, 62, 64, 65, 66, 68, 69, 71, 73, 75, 79, 80, 81, 83, 84, 86, 88, 90, 91, 92, 94, 96, 97, 98, 99
& 9, 14, 29, 39, 55, 57, 71, 85, 91, 95 \\

layers.1.CA
& 0, 10, 21, 34, 43, 53, 65, 81, 86, 94
& 13, 17, 36, 63, 65, 74, 88, 94 \\

layers.1.FFN
& 0, 4, 10, 19, 29, 42, 53, 64, 81, 90
& 2, 13, 25, 36, 56, 65, 72, 78, 89 \\

layers.2.SA
& 0, 5, 10, 20, 29, 42, 54, 69, 81, 93
& 6, 18, 34, 43, 59, 67, 81, 93 \\

layers.2.CA
& 0, 6, 11, 18, 28, 40, 50, 63, 80, 90
& 22, 41, 43, 45, 50, 70, 76, 77, 78, 85, 89 \\

layers.2.FFN
& 0, 5, 16, 28, 43, 56, 69, 81, 92, 97
& 9, 13, 22, 28, 39, 45, 51, 54, 63, 73, 76, 83, 97 \\

layers.3.SA
& 0, 12, 29, 41, 53, 69, 80, 89, 96, 98
& 9, 17, 34, 73, 74, 82, 83 \\

layers.3.CA
& 0, 6, 20, 30, 40, 52, 68, 80, 88, 96
& 6, 15, 38, 41, 48, 79, 81, 85, 93 \\

layers.3.FFN
& 0, 3, 11, 28, 43, 68, 80, 86, 94, 97
& 21, 27, 33, 36, 50, 54, 68, 74, 77, 79, 82 \\

layers.4.SA
& 0, 11, 26, 40, 55, 69, 81, 89, 95, 98
& 1, 8, 16, 17, 27, 40, 43, 46, 69, 79, 85, 90 \\

layers.4.CA
& 0, 4, 16, 21, 29, 39, 49, 58, 77, 91
& 3, 8, 21, 24, 38, 39, 41, 45, 48, 50, 54, 67, 94 \\

layers.4.FFN
& 0, 8, 29, 46, 62, 73, 84, 88, 92, 97
& 0, 16, 26, 50, 52, 67, 74, 84, 88, 92 \\

layers.5.SA
& 0, 19, 42, 61, 74, 83, 89, 93, 96, 98
& 43, 46, 72, 77, 83, 89, 91, 92 \\

layers.5.CA
& 0, 11, 28, 36, 47, 59, 74, 81, 88, 94
& 10, 22, 40, 46, 51, 53, 59, 72 \\

layers.5.FFN
& 0, 19, 36, 42, 43, 59, 64, 66, 71, 75, 79, 80, 83, 84, 88, 91, 92, 94, 96, 97, 98, 99
& 4, 6, 8, 9, 21, 32, 53, 54, 78, 85, 89 \\

layers.6.SA
& 0, 42, 67, 81, 87, 91, 93, 95, 97, 99
& 2, 8, 10, 14, 15, 43, 47, 54, 57, 64, 70, 77, 84, 96 \\

layers.6.CA
& 0, 6, 19, 42, 62, 75, 81, 86, 91, 95
& 4, 19, 33, 35, 42, 54, 66, 72, 75, 78, 79, 83, 88, 96 \\

layers.6.FFN
& 0, 19, 42, 43, 64, 66, 71, 75, 79, 80, 83, 88, 92, 94, 96, 97, 98
& 14, 23, 29, 41, 49, 65, 69, 70, 76, 84, 88 \\

layers.7.SA
& 0, 23, 46, 67, 80, 87, 91, 93, 95, 97
& 5, 22, 24, 41, 54, 55, 58, 66, 68, 77, 84, 85, 86 \\

layers.7.CA
& 0, 6, 17, 38, 64, 79, 86, 91, 94, 97
& 4, 20, 41, 48, 71, 72, 81 \\

layers.7.FFN
& 0, 42, 64, 71, 75, 79, 83, 88, 94, 97
& 5, 20, 22, 27, 30, 45, 63, 93, 99 \\

\end{longtable}

\subsection{Search Trajectories under Target-conditioned Early Stopping}
We further visualize the target-conditioned early-stopping process on two
representative tasks: Can-PH and Square-MH. The trajectories record
the best quick-evaluation score during evolutionary search, together with the
triggered formal evaluations and the final search termination point.

\begin{figure}[t]
    \centering

    \includegraphics[
        width=0.96\linewidth
    ]{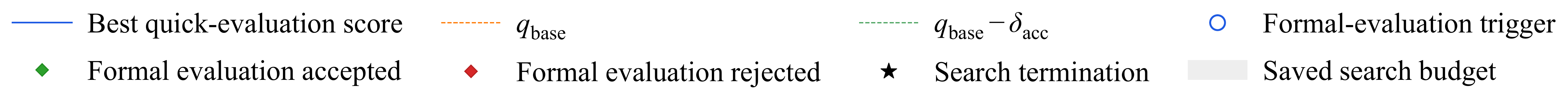}

    \vspace{1mm}

    \begin{minipage}[t]{0.49\linewidth}
        \centering
        \includegraphics[
            width=\linewidth
        ]{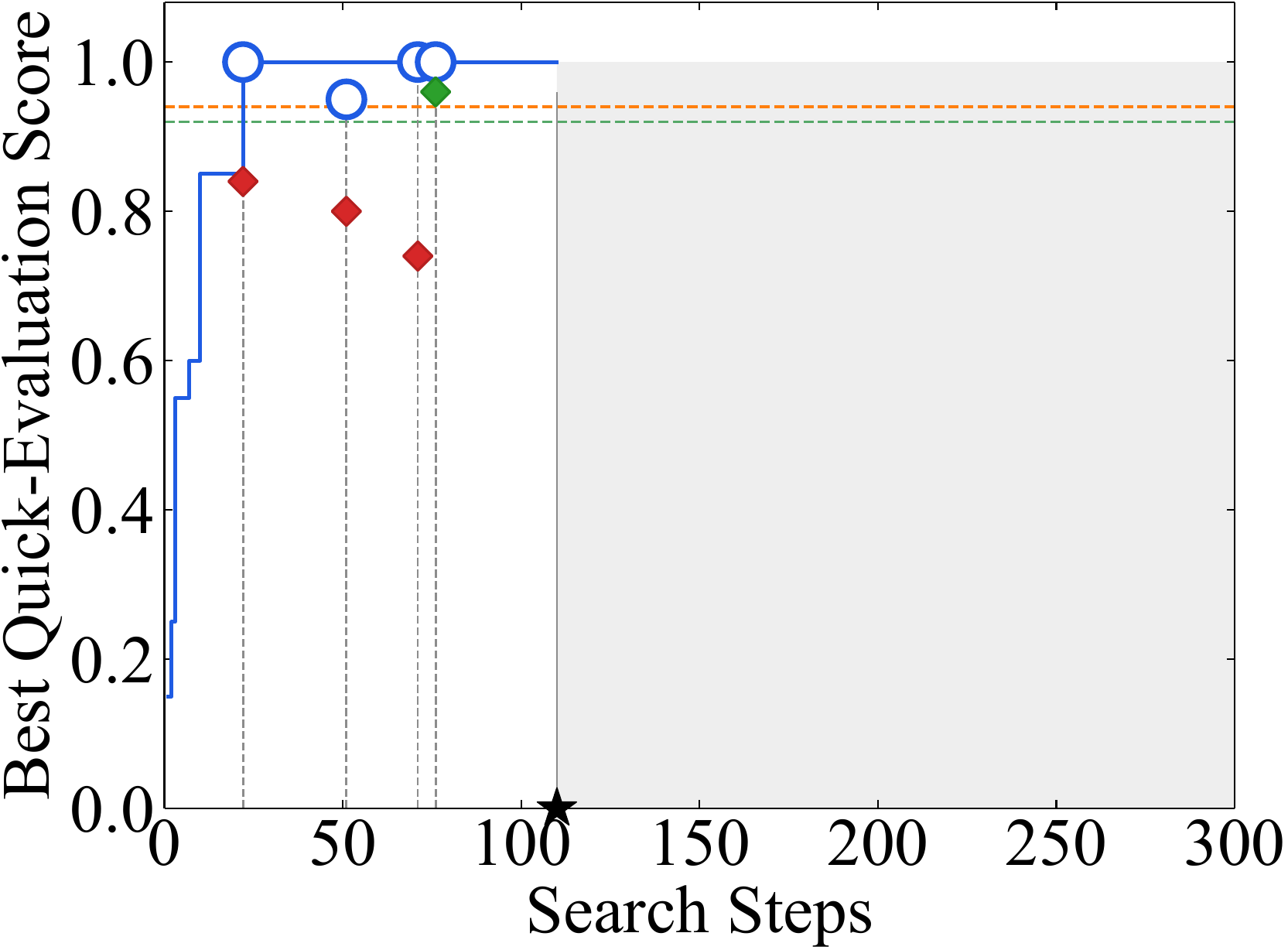}

        \vspace{-1mm}
        {\small (a) Can-PH}
    \end{minipage}
    \hfill
    \begin{minipage}[t]{0.49\linewidth}
        \centering
        \includegraphics[
            width=\linewidth
        ]{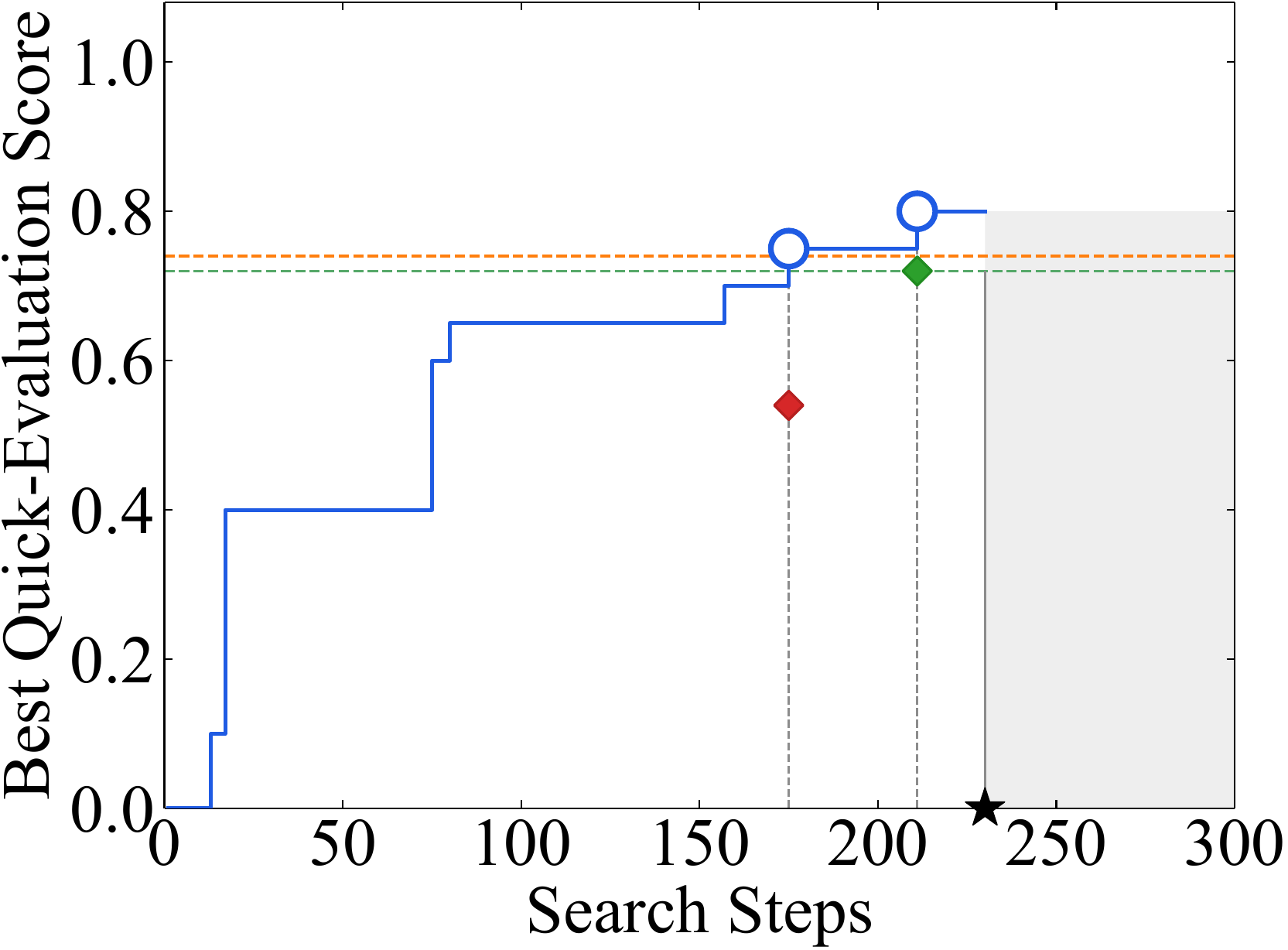}

        \vspace{-1mm}
        {\small (b) Square-MH}
    \end{minipage}

    \caption{Search trajectories under target-conditioned early stopping on
    representative tasks. EVO terminates the search once a candidate schedule
    passes formal evaluation, while a rejected formal evaluation allows the
    search to continue.}
    \label{fig:early-stopping-trajectories}
\end{figure}

As shown in Fig.~\ref{fig:early-stopping-trajectories}, EVO terminates the search once a
candidate passes formal evaluations, leaving part of the maximum search budget
unused. When formal evaluations fails, the search continues until another
candidate satisfies the acceptance criterion. These trajectories demonstrate
that the proposed mechanism reduces unnecessary evaluations while avoiding
premature termination caused by noisy quick-evaluation results.

\subsection{Runtime Analysis on Representative Tasks}

To complement the FLOPs-based analysis, we measure the wall-clock inference
time of a full policy inference call on five representative tasks. Timing begins
immediately before calling \texttt{policy.predict\_action} with observations
already resident on the GPU and ends when the call returns the predicted action
tensor. The measurement therefore covers the full execution of
\texttt{policy.predict\_action}, including visual encoding, diffusion sampling,
cache operations, and action post-processing.
\begin{table}[H]
    \centering
    \caption{Wall-clock inference time on representative tasks.}
    \label{tab:inference_time}
    \setlength{\tabcolsep}{4pt}

    \resizebox{\textwidth}{!}{
    \begin{tabular}{lcccccc}
        \toprule
        \multirow{2}{*}{Method}
        & \multicolumn{5}{c}{Inference Time (ms) $\downarrow$}
        & \multirow{2}{*}{AVG} \\
        \cmidrule(lr){2-6}
        & Transport PH
        & Transport MH
        & Kitchen
        & Tool PH
        & Square PH
        & \\
        \midrule

        Full Precision
        & 446.77
        & 440.09
        & 466.19
        & 437.55
        & 453.97
        & 448.91 \\
        \midrule

        EfficientVLA ($M=8$)
        & 124.16
        & 130.79
        & 118.29
        & 119.99
        & 115.52
        & 121.75 \\

        EfficientVLA ($M=10$)
        & 118.98
        & 133.50
        & 126.17
        & 111.55
        & 117.65
        & 121.57 \\
        \midrule

        BAC ($M=8$)
        & 153.96
        & 148.85
        & 144.46
        & 144.09
        & 138.51
        & 145.97 \\

        BAC ($M=10$)
        & 163.13
        & 152.64
        & 154.10
        & 146.56
        & 153.10
        & 153.90 \\
        \midrule

        \rowcolor{yellow!12}
        EVO ($M=8$)
        & \textbf{132.87}
        & 131.30
        & \textbf{129.27}
        & \textbf{134.32}
        & \textbf{133.07}
        & \textbf{132.17} \\

        \rowcolor{yellow!12}
        EVO ($M=10$)
        & 143.50
        & \textbf{120.46}
        & 144.37
        & 137.63
        & 145.20
        & 138.23 \\
        \bottomrule
    \end{tabular}
    }
\end{table}

All methods are evaluated on the same NVIDIA GeForce RTX 4090D GPU with a
batch size of 1. For each setting, we conduct three independent runs
sequentially on this GPU, each in a fresh process. Each run consists of 10
warm-up calls followed by 50 timed calls, yielding 150 measurements in total.
CUDA execution is synchronized immediately before and after each timed call,
and the mean inference time over all measurements is reported.

As shown in Table~\ref{tab:inference_time}, EVO reduces the average inference
time of Full Precision from 448.91 ms to 132.17 ms and 138.23 ms for $M=8$
and $M=10$, respectively. The corresponding policy inference frequencies,
computed as the reciprocals of the average inference time, increase from
2.23 Hz to 7.57 Hz and 7.23 Hz. Although EfficientVLA achieves a lower
inference time, the main results show that it incurs greater degradation in task
performance. Taken together, these results indicate that EVO achieves a more
favorable trade-off between inference efficiency and task performance.